\documentclass[pdflatex,sn-mathphys-num]{sn-jnl}%
\usepackage{fix-cm}
\usepackage{graphicx}%
\usepackage{subcaption} 
\usepackage{multirow}%
\usepackage{amsmath,amssymb,amsfonts}%
\usepackage{amsthm}%
\usepackage{mathrsfs}%
\usepackage[title]{appendix}%
\usepackage{xcolor}%
\usepackage{textcomp}%
\usepackage{manyfoot}%
\usepackage{booktabs}%
\usepackage{algorithm}%
\usepackage{algorithmicx}%
\usepackage{algpseudocode}%
\usepackage{listings}%
\usepackage{xr}
\usepackage{makecell}
\usepackage{caption}
\usepackage{amsmath} 
\usepackage{bm}

\begin{document}
\title[Article Title]{MolGraph-xLSTM: A graph-based dual-level xLSTM framework with multi-head mixture-of-experts for enhanced molecular representation and interpretability}

\author[1,2]{\fnm{Yan} \sur{Sun}}\email{ysun2443@uwo.ca}
\author[3]{\fnm{Yutong} \sur{Lu}}\email{lorraine.lu@mail.utoronto.ca}
\author[3]{\fnm{Yan Yi} \sur{Li}}\email{yanyi.li@mail.utoronto.ca}
\author[2]{\fnm{Zihao} \sur{Jing}}\email{zjing29@uwo.ca}
\author[1]{\fnm{Carson K.} \sur{Leung}}\email{carson.leung@umanitoba.ca}
\author*[1,2,3,4,5,6]{\fnm{Pingzhao} \sur{Hu}}\email{phu49@uwo.ca}

\affil[1]{\orgdiv{Department of Computer Science}, \orgname{University of Manitoba}, \orgaddress{\street{66 Chancellors Cir}, \city{Winnipeg}, \postcode{R3T 2N2}, \state{Manitoba}, \country{Canada}}}

\affil[2]{\orgdiv{Department of Computer Science}, \orgname{Western University}, \orgaddress{\street{1151 Richmond St}, \city{London}, \postcode{N6A 3K7}, \state{Ontario}, \country{Canada}}}
\affil[3]{\orgdiv{Biostatistics Division, Dalla Lana School of Public Health}, \orgname{University of Toronto}, \orgaddress{\street{155 College St}, \city{Toronto}, \postcode{M5T 3M7}, \state{Ontario}, \country{Canada}}}

\affil[4]{\orgdiv{Department of Biochemistry}, \orgname{Western University}, \orgaddress{\street{1151 Richmond St}, \city{London}, \postcode{N6A 3K7}, \state{Ontario}, \country{Canada}}}

\affil[5]{\orgdiv{Department of Oncology}, \orgname{Western University}, \orgaddress{\street{1151 Richmond St}, \city{London}, \postcode{N6A 3K7}, \state{Ontario}, \country{Canada}}}

\affil[6]{\orgdiv{Department of Epidemiology and Biostatistics}, \orgname{Western University}, \orgaddress{\street{1151 Richmond St}, \city{London}, \postcode{N6A 3K7}, \state{Ontario}, \country{Canada}}}

\abstract{Predicting molecular properties is essential for drug discovery, and computational methods can greatly enhance this process. Molecular graphs have become a focus for representation learning, with Graph Neural Networks (GNNs) widely used. However, GNNs often struggle with capturing long-range dependencies. To address this, we propose MolGraph-xLSTM, a novel graph-based xLSTM model that enhances feature extraction and effectively models molecule long-range interactions.

Our approach processes molecular graphs at two scales: atom-level and motif-level. For atom-level graphs, a GNN-based xLSTM framework with jumping knowledge extracts local features and aggregates multilayer information to capture both local and global patterns effectively. Motif-level graphs provide complementary structural information for a broader molecular view. Embeddings from both scales are refined via a multi-head mixture of experts (MHMoE), further enhancing expressiveness and performance.

We validate MolGraph-xLSTM on 10 molecular property prediction datasets, covering both classification and regression tasks. Our model demonstrates consistent performance across all datasets, with improvements of up to $7.03\%$ on the BBBP dataset for classification and $7.54\%$ on the ESOL dataset for regression compared to baselines. On average, MolGraph-xLSTM achieves an AUROC improvement of $3.18\%$ for classification tasks and an RMSE reduction of $3.83\%$ across regression datasets compared to the baseline methods. These results confirm the effectiveness of our model, offering a promising solution for molecular representation learning for drug discovery.}

\keywords{molecular property prediction, molecular graph representation learning, multi-head mixture-of-experts, drug discovery, xLSTM}

\maketitle
\section{Introduction}
Predicting the molecular properties of a compound, particularly its ADMET (Absorption, Distribution, Metabolism, Excretion, and Toxicity) characteristics, is critical during the early stages of drug development \citep{2024mldd,Jia2022}. Leveraging deep learning for molecular representation to predict their properties significantly enhances the efficiency of identifying potential drug candidates \citep{2020nmidrug,2023naturedrug}. Molecular graphs retain richer structural information, which is crucial for accurate property prediction. In recent years, graph neural networks (GNNs) built on molecular graph data have been extensively utilized for molecular representation learning to predict their properties \citep{2019attentivefp,2019DMPNN,deepergcn2020,GROVER,2022moclr,fpgnn2022,2022GEM,2023transfoxmol,hierachicalmolgraphselfsupervised2023}.

A key challenge in molecular property prediction lies in capturing long-range dependencies—the influence of distant atoms or substructures within a molecule on a target property. While graph neural networks (GNNs) leverage neighborhood aggregation as their core mechanism—updating the hidden states of each node by aggregating information from neighboring nodes using operations like sum, max, or mean pooling \citep{MPNN2020,2022gnnreview}—they face significant limitations in capturing these long-range dependencies. Specifically, over-smoothing and over-squashing hinder their performance. Over-smoothing occurs when, as the number of layers increases, node representations become increasingly similar, leading to a loss of distinction between nodes \citep{2018DeeperII}. On the other hand, over-squashing refers to the compression of information from distant nodes as it propagates toward the target node, making it challenging for relevant information to be effectively transmitted \citep{Alon2020OnTB}. These issues limit the ability of GNNs to fully exploit global structural information, reducing their effectiveness in complex molecular property prediction tasks.

To address these challenges, we propose the MolGraph-xLSTM model, which integrates the xLSTM architecture with molecular graphs. Traditionally, Long Short-Term Memory (LSTM) networks have been widely applied in natural language processing (NLP) tasks to capture sequential data representations \citep{Hochreiter1997LSTM}. With its gating mechanisms, the LSTM can effectively decide which information to retain or discard, allowing it to manage long-range dependencies. Thus, we incorporate LSTM into our model to address the limitations of GNNs in handling long-range information. Recently, an improved version of LSTM, called xLSTM, was introduced \citep{Beck2024xLSTM}. The xLSTM includes two additional modules, sLSTM and mLSTM, which expand the storage capacity of the original LSTM. Experimental results for xLSTM have shown favorable performance compared to two state-of-the-art architectures: Transformer \citep{2017Transformer} and State Space Models \citep{2023mamba}. For this reason, we choose this enhanced LSTM model in our framework.

\begin{figure}[htbp]
  \centering
  \begin{subfigure}[b]{0.45\textwidth}
    \centering
    \includegraphics[width=\textwidth]{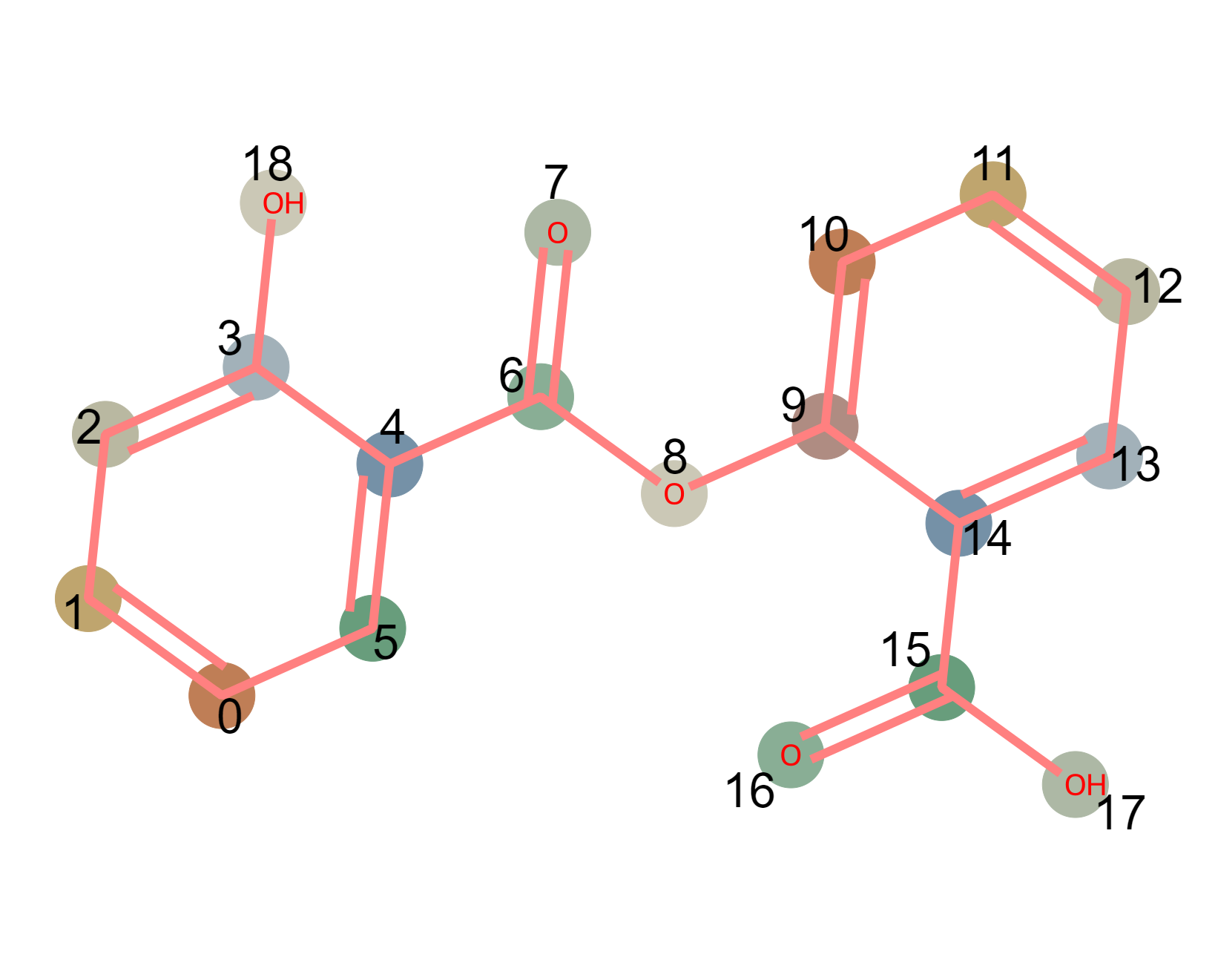}
    \caption{atom-level graph}
  \end{subfigure}
  \hfill
  \begin{subfigure}[b]{0.45\textwidth}
    \centering
    \includegraphics[width=\textwidth]{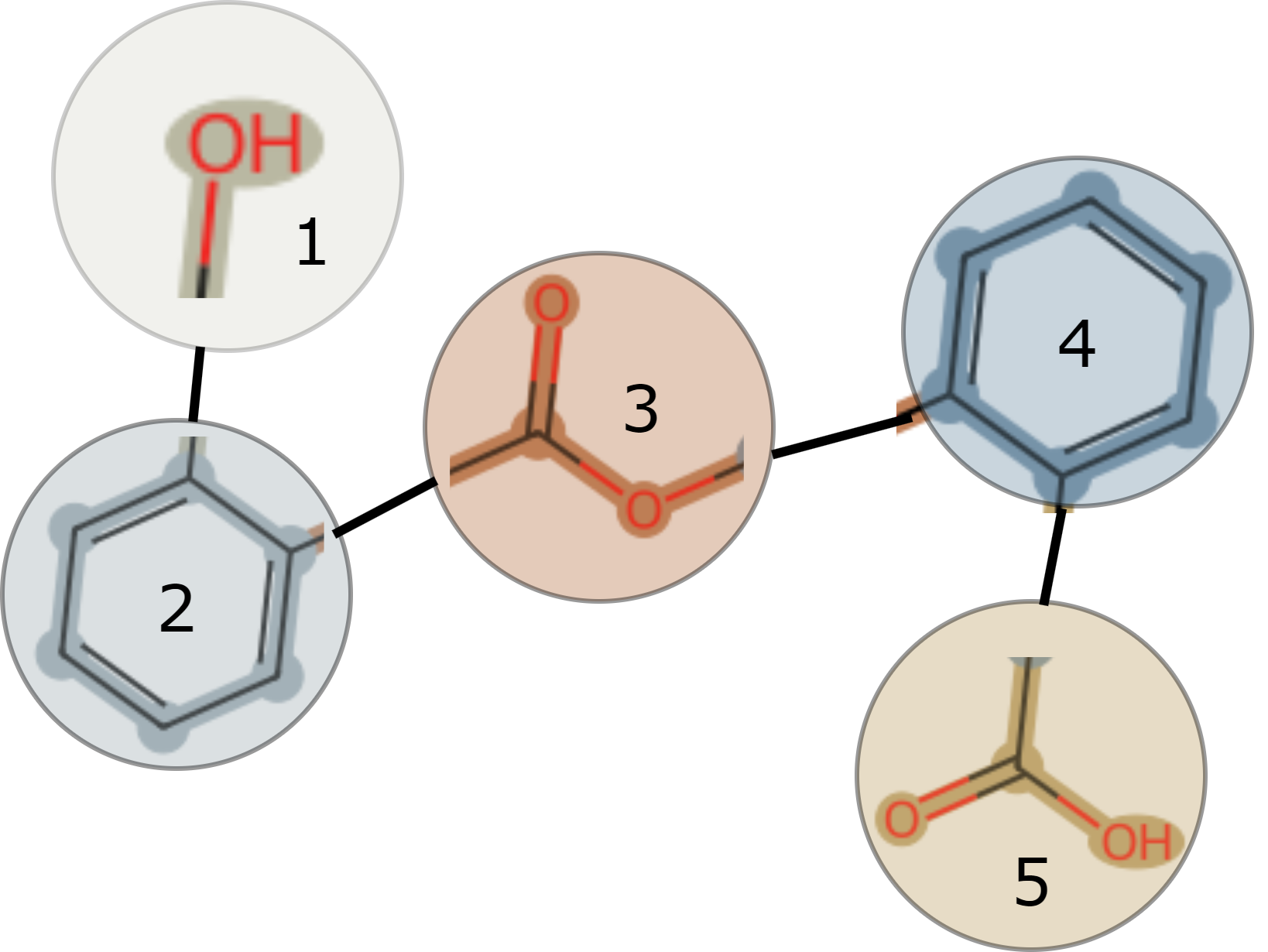}
    \caption{motif-level graph}
  \end{subfigure}
  \caption{\textbf{Comparison of atom-level and motif-level graph representations.}
(a) The atom-level graph represents each atom as a node and each bond as an edge. (b) The motif-level graph combines substructures into single nodes, resulting in a graph that is less complex than the atom-level graph.}
  \label{figure:atomvsfg}
\end{figure}

We utilize both atom-level and motif-level molecular graphs in our approach (Figure \ref{figure:atomvsfg}). In the atom-level graph, each node represents an atom and each edge represents a bond within the molecule. The motif-level graph, on the other hand, is a partitioned version of the atom-level graph, where each node represents a substructure (such as aromatic ring) within a molecule. This results in a significantly simplified representation compared to the atom-level graph. Such simplification aids the model in learning features linked to local structures, as similar local motifs, from a functional group perspective, tend to impart similar properties to molecules \citep{2017FG}. Furthermore, the simplified motif-level graph, by reducing complexity and eliminating cycle structures, becomes more closely with sequential data. This structural simplification aligns well with the strengths of xLSTM, which is inherently designed to handle sequential information, making the motif-level graph more suitable for processing with xLSTM.

However, relying solely on the motif-level graph would not capture all molecular details effectively, and motif partitioning itself demands precise segmentation. Therefore, we incorporate both atom-level and motif-level graphs in our model. For the atom-level representation, we introduce a GNN-based xLSTM with jumping knowledge \citep{Xu2018JK}. Here, the GNN collects local information from the atom-level graph, and jumping knowledge aggregates features from multiple GNN layers, producing enriched node representations as inputs to xLSTM. By combining features from both the atom- and motif-level graphs, we constructed a comprehensive molecular representation for accurate property prediction. 

Additionally, we integrate the multi-head mixture-of-experts (MHMoE) module \citep{2024MMoE} to enhance the predictive performance of our model. The sparse mixture-of-experts (SMoE) \citep{2017SMoE} framework has been demonstrated as an effective method for scaling models while maintaining computational efficiency by dynamically assigning inputs to different expert networks. This allows the input features to be processed by multiple experts, enabling diverse perspectives and improving the quality of learned representations. Building upon SMoE, the MHMoE architecture introduces further advancements by enhancing the usage of expert and promoting a more fine-grained understanding of input features. By incorporating the MHMoE module, our model is able to generate more expressive feature representations, which enhances its predictive accuracy.

The contributions of our work are as follows:

\begin{itemize}
    \item \textbf{Development of a dual-level molecular graph representation framework}:  
    We developed a novel representation learning framework for property prediction that leverages both atom-level and motif-level molecular graph representations. This dual-level approach captures fine-grained molecular details and higher-level structural features, demonstrating its effectiveness across 10 molecular property prediction datasets.
    
    \item \textbf{Adaptation of xLSTM to molecular graphs}:  
    We introduced the advanced xLSTM architecture into molecular graph representation learning, addressing the limitations of traditional GNNs in capturing long-range dependencies. Our approach achieved improved performance in molecular property prediction tasks compared to four baseline models.
    
    \item \textbf{Integration of Multi-Head Mixture-of-Experts (MHMoE) for enhanced prediction}:  
    We incorporated the multi-head mixture-of-experts (MHMoE) module into our framework, which dynamically assigns input features to different expert networks, enabling diverse feature processing and improving predictive accuracy. This architecture refines feature representations through fine-grained expert activation.
    
    \item \textbf{Case study analysis for model interpretability}:  
    We conducted case study to investigate the substructures assigned the highest weights by the network, demonstrating that the atom-level and motif-level information are complementary. By cross-referencing with known literature, we identified strong correlations between the highlighted substructures and specific molecular properties, underscoring the ability of the model to implicitly learn biologically relevant information.
\end{itemize}

\section{Background and Related Work} \label{sec2}
\subsection{Extended Long Short-Term Memory} \label{section2.1}
Long Short-Term Memory (LSTM) networks are designed to process sequence data by incorporating a memory cell regulated by three gates: the input gate $i_t$, output gate $o_t$, and forget gate $f_t$. The input gate controls how much new information is added to the memory, the forget gate decides how much of the past information to retain, and the output gate determines what part of the memory contributes to the current hidden state. At each time step $t$, the memory cell is updated by combining the retained memory from the previous step with new information, ensuring the network can selectively remember or forget information as needed. The update of the memory cell can be represented as:
\begin{align}
    c_t &= f_t \odot c_{t-1} + i_t \odot z_t, \label{cellstate}\\
    h_t &= o_t \odot \psi (c_t),  \\
    z_t &= \varphi (w_z^\top x_t + r_z h_{t-1} + b_z),\\
    i_t &= \sigma (w_i^\top x_t + r_i h_{t-1} + b_i), \label{inputgate}\\
    f_t &= \sigma (w_f^\top x_t + r_f h_{t-1} + b_f),\label{forgetgate}\\
    o_t &= \sigma (w_o^\top x_t + r_o h_{t-1} + b_o), 
\end{align}
where $c_t$ denotes the cell state, $h_t$ represents the hidden state and $z_t$ represents the candidate state. The terms $w_z$, $w_i$, $w_f$, and $w_o$ are weight vectors associated with the input vector $x_t$, while $r_z$, $r_i$, $r_f$, and $r_o$ are weight vectors corresponding to the previous hidden state $h_{t-1}$. The bias terms are given by $b_z$, $b_i$, $b_f$, and $b_o$. The functions $\psi(\cdot)$, $\varphi(\cdot)$, and $\sigma(\cdot)$ are activation functions, where $\psi(\cdot)$ and $\varphi(\cdot)$ typically represent $tanh$ function, and $\sigma(\cdot)$ denotes the $sigmoid$ function.

In xLSTM, two new blocks are introduced: sLSTM and mLSTM. Both incorporate exponential gating to enhance the memory cells of the original LSTM. The function $\sigma(\cdot)$ in equations \ref{inputgate} and \ref{forgetgate} is modified to use exponential activation. mLSTM further extends the memory capacity by introducing a matrix memory, which improves the storage capabilities of LSTM. Specifically, it replaces the scalar cell state $c$ in equation \ref{cellstate} with a matrix. The xLSTM block is constructed by stacking alternating sLSTM and mLSTM blocks, while the full xLSTM architecture consists of multiple xLSTM blocks stacked together. \\

\subsection{Deep Learning Methods Based on Molecular Graphs for representation learning to predict molecular properties}
GNN is the most widely used architecture for molecular graph representation. GNN updates node features through a two-step process: first, aggregating information from neighboring nodes, followed by applying a Multi-Layer Perceptron (MLP) to update the own features. Various GNN-based models have been proposed for molecular property prediction. The Directed-Message Passing Neural Network (DMPNN) \citep{2019DMPNN} optimizes message passing by centering aggregation on bonds rather than atoms, effectively avoiding redundant loops. DeeperGCN \citep{deepergcn2020} focuses on training very deep GCNs by introducing an improved residual connection to enhance performance. The Hierarchical Informative Graph Neural Network (HiGNN) \citep{HiGNN2022} segments the molecular graph into fragments using Breaking of Retrosynthetically Interesting Chemical Substructures (BRICS) \citep{2008BRICS}. Both the original molecular graph and its fragments are then processed by the GNN to generate a hierarchical representation of the molecule.  FP-GNN \citep{fpgnn2022} integrates molecular fingerprints with Graph Attention Networks (GAT), combining traditional descriptors with GNN features to improve prediction accuracy. These approaches primarily rely on GNNs for feature extraction. While DeeperGCN and DMPNN focus on optimizing message-passing mechanisms, HiGNN introduces hierarchical information through fragment-based graph segmentation, and FP-GNN enhances GNN-based features by incorporating traditional molecular fingerprints. Beyond GNN-exclusive models, hybrid architectures have been proposed. For example, TransFoxMol \citep{2023transfoxmol} designed an embedding unit that combines a GNN and a transformer to balance the local and long-range interactions of an atom.

\subsection{Multi-Head Mixture of Experts} \label{sec2moe}
The Mixture of Experts (MoE) is a classical ensemble method that combines multiple experts with identical architectures, routing inputs to specific experts via a gating mechanism \citep{1991AMoE, 2022TUMoE}. This design enables different experts to specialize in processing distinct types of information.

Recently, the application of MoE in deep learning has gained significant attention. The Sparsely-Gated Mixture-of-Experts (SMoE) layer was introduced, where each input is routed to the top-K ($K \geq 2$) most appropriate experts \citep{2017SMoE}. Building upon this, the Multi-Head Mixture-of-Experts (MHMoE) \citep{2024MMoE} was proposed, which partitions the input into multiple segments, enabling each segment to be processed by the top-K selected experts.

Consider a system with $n$ experts, denoted as $E_1, E_2, \dots, E_n$, and an input $x \in \mathbb{R}^{h \times d}$. The input $x$ is divided into $h$ segments, $x_1, x_2, \dots, x_h$, where each segment is $d$-dimensional. The output of the MoE layer for a given segment $x_s$ is calculated as:

\begin{align}
    x_s^{\text{MoE}} = \sum_{e=1}^{n} G(x_s)_e E_e(x_s), \label{moeEquation}
\end{align}

where $G(x_s)_e$ is the gating function that assigns a weight to the $i$-th expert. The gating function $G(x_s)_e$ is defined as:

\begin{align}
G(x_s) = \text{softmax}(\text{TopK}(g(x_s) + D_{\text{noise}}, K)),
\end{align}

with $g(x_s)$ computing the raw scores for each expert and $D_{\text{noise}}$ adding stability and exploration during training. The TopK function filters the top-$K$ elements of a vector $v$ as follows:

\begin{align}
\text{TopK}(v, K)_s = 
\begin{cases} 
v_s & \text{if } v_s \text{ is among the top } K \text{ elements of } v, \\
-\infty & \text{otherwise.}
\end{cases}
\end{align}

Finally, the MHMoE output is obtained by concatenating the outputs of all segments processed by the MoE layer:

\begin{align}
x^{\text{MHMoE}} = CONCAT(x_1^{\text{MoE}}, x_2^{\text{MoE}}, \dots, x_h^{\text{MoE}}).
\end{align}

\section{Method}
\begin{figure}[!t]
\centering
\includegraphics[width=1\textwidth]{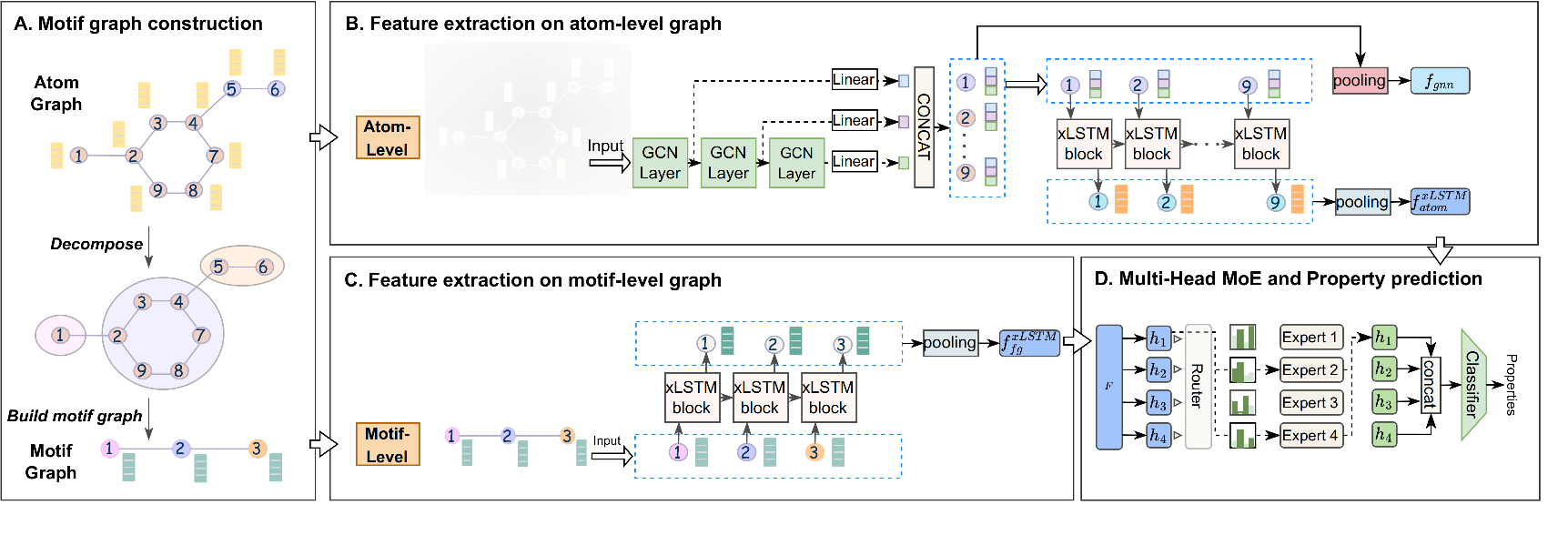}
\caption{
    \textbf{Architecture of MolGraph-xLSTM.} 
    The architecture consists of four main components: 
    (A) Motif graph construction. The atom-level graph is decomposed into motifs to form a motif-level graph. 
    (B) Feature extraction on the atom-level graph. A GCN-based xLSTM framework with jumping knowledge extracts features, followed by pooling to generate the atom-level representation $f_{atom}^{xLSTM}$. 
    (C) Feature extraction on the motif-level graph. Using xLSTM blocks and pooling to produce the motif-level representation $f_{motif}^{xLSTM}$. 
    (D) Multi-Head Mixture-of-Experts (MHMoE) and property prediction. Features ($f_{gcn}$, $f_{atom}^{xLSTM}$ and $f_{motif}^{xLSTM}$) are combined and refined through the MHMoE module for final property prediction.
    }
\label{archi}
\end{figure}

\subsection{Construction of Atom- and Motif-Level Molecular Graphs}
Starting from the SMILES string of a molecule, we convert it into an atom-level molecular graph $G_{\text{atom}} = \{V_{\text{atom}}, E_{\text{atom}}\}$ using RDKit tool \citep{2006rdkit}, where $V_{\text{atom}} = \{ v^{\text{atom}}_p \}$ represents the set of nodes, and $E_{\text{atom}} = \{ (v^{\text{atom}}_p, v^{\text{atom}}_q) \}$ represents the set of edges. Each node $v^{\text{atom}}_p$ corresponds to an atom and is initialized with 11 atomic features, including atomic number, chirality, and aromaticity (Table \ref{tab:atomfeatures}). Likewise, each edge $(v^{\text{atom}}_p, v^{\text{atom}}_q)$ represents a bond and includes features such as bond type, stereochemistry, and conjugation (Table \ref{tab:bondfeatures}). Based on the atom-level graph, we then generate a motif-level graph $G_{motif} = \{V_{motif}, E_{motif}\}$ through the ReLMole, as described by \citep{ReLMole}. In RelMole, three types of substructures are considered as motifs: rings, non-cyclic functional groups, and carbon-carbon single bonds. In this motif graph, each node represents a motif and is initilized with 12 features. Each edge represents the connection between two motifs. Details of the initial features are provided in Table \ref{tab:motif-graph-features} in supplementary information. 

Both node and edge features are embedded into a $d$-dimensional feature vector. Specifically, we denote the input node feature of the atom-level graph and the motif-level graph as $H_{atom}^0 \in \mathbb{R}^{N_{atom} \times d}$ and $H_{motif}^0 \in \mathbb{R}^{N_{motif} \times d}$, respectively. $N_{atom}$ represents the number of atoms and $N_{motif}$ is the number of motif. The input feature of the edge in the atom-level graph between nodes $p$ and $q$ is $e_{pq} \in \mathbb{R}^d$.

\subsection{Feature Extraction on Atom-Level Graph} \label{atom-level}
\subsubsection{Graph Neural Network} \label{GNN}
In the graph neural network (GNN) component, we employ a simplified message-passing mechanism that incorporates both residual connections \cite{deepergcn2020} and virtual nodes \cite{MPNN2020}. At each GNN layer, the process starts by applying layer normalization ($LN$) to the node representations, followed by a $ReLU$ activation. To facilitate the exchange of global information across the graph, we introduce virtual nodes, which aggregate the features of all nodes in the graph. The resulting virtual node information is then added to the individual node representations. The operations can be formally expressed as:

\begin{align}
 &h_p^{l+1} = ReLU(LN(h_p^l)) + + vn^{l+1}, \\
 &vn^{l+1} = \sum_{k=1}^{N_{atom}} h_k^l,
\end{align}

where $h_p^{l+1} \in \mathbb{R}^d$ denotes the hidden state of node $p$ at layer $l+1$, $vn^{l+1}$ represents the vector of the virtual node.

Next, the message-passing step occurs, where the information from neighboring nodes and the edges connecting them is aggregated. For each edge $e_{pq}$, a message is computed as: $m_{pq} = h_q^{l+1} + e_{pq}$. The messages from all neighboring nodes $\mathcal{N}(p)$ are summed and used to update the node representation through a MLP: 
\begin{align}
h_p^{t+1} = MLP\Big(\sum_{q \in \mathcal{N}(p)}m_{pq}\Big).
\end{align}

Finally, a residual connection is applied, adding the original node representation from layer $l$ to the updated node representation at layer $l+1$: $h_p^{l+1} \gets h_p^{l+1} + h_p^l$.

\subsubsection{Jumping Knowledge} \label{jk}
After the GNN, we apply a jumping knowledge to aggregate information from all GNN layers. This allows each node feature to encapsulate representations from both shallow and deep layers. The operation is defined as:

\begin{align}
    h_p^{GNN} = \text{CONCAT}(h_p^1A_1^T, h_p^2A_2^T, \ldots, h_p^lA_l^T),
\end{align}

where $h_p^{GNN} \in \mathbb{R}^{d_{skip} \times num_{jk}}$ represents the aggregated feature of node $p$ from the GNN, and $A_l^T \in \mathbb{R}^{d \times d_{skip}}$ is a weight matrix that maps the layer-specific node feature $h_p^l \in \mathbb{R}^d$ to a lower-dimensional space. In our experiments, we evaluate the impact of the number of jumping knowledge layers $num_{jk}$ on performance. 

\subsubsection{Using xLSTM to Capture the Long-Range Information} \label{xLSTM}
In this section, we utilize xLSTM to capture long-range dependencies for each node in the graph. We treat the output of the GNN, $H^{GNN} \in \mathbb{R}^{N_{atom} \times (d_{skip} \times num_{jk})}$, as a sequence of length $N_{atom}$, where each node corresponds to one element of the sequence. This sequence is then passed through the xLSTM model, producing an output $H_{atom}^{xLSTM} \in \mathbb{R}^{N_{atom} \times (d_{skip} \times num_{jk})}$, as follows:
\begin{align}
    H_{atom}^{xLSTM} = xLSTM(H^{GNN}).
\end{align}

\subsection{Motif-Level Feature Extraction} \label{fg-level}
The motif-level graph is processed directly by the xLSTM model. We first map the input feature $H_{motif}^0$ to the dimension $d_{skip} \times num_{jk}$, matching the output dimension of the atom-level graph. This mapped feature is then passed through the xLSTM model to produce an output $H_{motif}^{xLSTM} \in \mathbb{R}^{N_{motif} \times (d_{skip} \times num_{jk})}$:

\begin{align}
    H_{motif}^{xLSTM} = xLSTM(H_{motif}^0).
\end{align}

\subsection{Perform a Multi-Head Mixture-of-Experts on the Features} \label{MHMoE}
We apply a global pooling operation to $H_{GNN}$, $H_{atom}^{xLSTM}$, and $H_{motif}^{xLSTM}$ to produce three graph-level features: $f_{GNN}$, $f_{atom}^{xLSTM}$, and $f_{motif}^{xLSTM}$. The final molecular feature, $f_{out}$, is obtained by summing these three features. Subsequently, we utilize a  Multi-Head Mixture-of-Experts (MHMoE), as described in Section \ref{sec2moe}, to process these features. For any input feature $f$, we split it into $h_{moe}$ segments and the output of the MHMoE for each segment $f_s$ is expressed as:
\begin{align}
    f_s^{MoE} = \sum_{i=1}^{n} G((f_s)_e E_e(f_s). 
\end{align}
In our work, each expert $E_e$ is a feedforward network (FFN) consisting of a variable number of stacked fully connected layers with activation functions between them. The number of stacked layers is a hyperparameter. The final output of the MHMoE module is the concatenation of the output of all segments: $f^{MHMoE} = CONCAT(f_1^{MoE}, f_2^{MoE}, ..., f_{h_{moe}}^{MoE})$. 

\subsection{Overall architecture}
The overall architecture is illustrated in Figure \ref{archi}. We perform feature extraction on both the atom-level graph and the motif-level graph. For the atom-level graph, we first apply the GNN (Section \ref{GNN}), followed by a skip connection (Section \ref{jk}) that aggregates the outputs from all GNN layers, resulting in $H^{\text{GNN}}$. This aggregated output is then passed through the xLSTM module (Section \ref{xLSTM}), producing $H_{\text{atom}}^{\text{xLSTM}}$. Next, global pooling is applied separately to $H^{\text{GNN}}$ and $H_{\text{atom}}^{\text{xLSTM}}$ to obtain global features of the graph from the GNN ($f_{\text{GNN}}$) and from the xLSTM ($f_{\text{atom}}^{\text{xLSTM}}$). These two features are then summed to generate $f_{\text{atom}} \in \mathbb{R}^{d_{\text{skip}} \times num_{jk}}$, representing the feature of the atom-level graph.

The motif-level graph is fed directly into the xLSTM model, yielding $H_{{motif}}^{{xLSTM}}$ (Section \ref{fg-level}). We obtain a feature $f_{{motif}} \in \mathbb{R}^{d_{{skip}} \times num_{jk}}$ for the motif-level graph by applying global pooling on $H_{{motif}}^{{xLSTM}}$. And then, $f_{{atom}}$ and $f_{{motif}}$ are summed to form the final feature of the molecule, which is then passed through an MHMoE module (section \ref{MHMoE}) to further improve the feature representation. Finally, the final representation is passed through MLP module to predict the molecular property:
\begin{align}
&f_{out} = MHMoE(f_{atom} + f_{motif}), \\
&output = MLP(f_{out}),
\end{align}
where $output \in \mathbb{R}^{K}$, and $K$ represents the number of tasks.

\subsection{Loss function}
To optimize the model, we applied two losses: the task loss and the supervised contrastive loss \citep{2020SCL}. The task loss is intended to guide the model in minimizing the error between the true label and the predicted value, while the supervised contrastive loss optimizes the feature embedding space by encouraging samples with the same label to be close to each other in the embedding space and separating those with different labels.

\subsubsection{Task loss}
For classification tasks, we use the cross-entropy loss, which measures the difference between the true label $y_i$ and the predicted probability distribution $\hat{y}_i$. This loss is formulated as:

\begin{align}
   \mathcal{L}_{task}^{classification} = - \sum_{k=1}^{K} y_{i,k} \log(\hat{y}_{i,k}), 
\end{align}

where $y_{i,k}$ represents true label and $\hat{y}_{i,k}$ is the predicted probability for task $k$.

For regression tasks, we adopt the mean squared error (MSE) loss, which captures the discrepancy between the predicted value $\hat{y}_i$ and the true value $y_i$. The MSE loss is expressed as:

\begin{align}
    \mathcal{L}_{task}^{regression}  = (y_{i} - \hat{y}_{i})^2.
\end{align}

\subsubsection{Supervised contrastive loss for classification task}
We apply the supervised contrastive loss (SCL) to all features: $f_{out}$, $f_{atom}$, and $f_{motif}$. To illustrate the process, we describe the calculation using $f_{out}$. First, we normalize $f_{out}$ as follows:
\begin{align}
    f_{out}^{norm} &= \frac{f_{out}}{\|f_{out}\|_2 + \epsilon}, \\
    \|f_{out}\|_2 &= \sqrt{\sum_{d} f_{out_d}^2},
\end{align}
where $\epsilon$ is a small constant added to prevent numerical instability, and $d$ indexes the dimensions of the feature vector.

Next, we compute the supervised contrastive loss $\mathcal{L}_{SCL}$ using the normalized feature $f_{out}^{norm}$:
\begin{align}
    \mathcal{L}_{SCL} = \sum_{i \in I} \frac{-1}{|P(i)|} \sum_{p \in P(i)} 
    \log \frac{\exp(f^{norm}_{out_i} \cdot f^{norm}_{out_p} / \tau)}{\sum_{a \in A(i)} \exp(f^{norm}_{out_i} \cdot f^{norm}_{out_a} / \tau)}.
\end{align}
In this equation, $i$ denotes the index of the anchor molecule. The set $P(i)$ includes indices of all samples sharing the same label as the anchor molecule, while $A(i)$ represents the set of all sample indices excluding $i$. $\tau$ is a temperature parameter

\subsubsection{Supervised contrastive loss for regression task}
For the regression task, it is necessary to define positive samples. This is achieved by computing the Euclidean distance between the labels of all sample pairs in the training set. From these distances, the median value $d_{med}$ and the maximum value $d_{max}$ are obtained. A sample is considered as a positive sample for a given anchor if its distance to the anchor is less than $d_{med}$. Additionally, weights are assigned to all samples to reflect their relative importance. Samples in $P(i)$ that are closer to the anchor sample are given higher importance, while samples in $A(i)$ that are farther from the anchor sample are assigned greater importance. The loss function is formulated as follows:

\begin{align}
    \mathcal{L}_{SCL} = \sum_{i \in I} \frac{-1}{|P(i)|} \sum_{p \in P(i)} w_p 
    \log \frac{\exp(f^{norm}_{out_i} \cdot f^{norm}_{out_p} / \tau)}{\sum_{a \in A(i)} w_a\exp(f^{norm}_{out_i} \cdot f^{norm}_{out_a} / \tau)},
\end{align}

where the weights are defined as:
\begin{align}
    &w_p = \frac{(d_{med} - d_{ip})}{d_{med}}, \\
    &w_a = \exp\Big(\frac{d_{ia} - d_{med}}{d_{max} - d_{med}}\Big). 
\end{align}

Here, $d_{ip}$ and $d_{ia}$ denote the Euclidean distances between sample $i$ and sample $p$, and between sample $i$ and sample $a$, respectively.

\subsubsection{Overall Loss Function}
The total loss function for training is a combination of the task loss and the supervised contrastive loss, given as:
\begin{align}
    \mathcal{L}_{total} = \mathcal{L}_{task} + \mathcal{L}_{SCL}.
\end{align}

\section{Experiments}
\subsection{Datasets and evaluation}
MoleculeNet \citep{2018moleculenet} is a benchmark collection designed to evaluate models for molecular property prediction, comprising datasets for both classification and regression tasks. In addition to MoleculeNet, we include the Caco-2 dataset \citep{2016caco2} for regression tasks. For dataset splitting, we adopted different strategies based on the nature of the tasks. For single-task classification datasets, we employed scaffold splitting, which ensures that structurally distinct molecules are separated into training, validation, and test sets. This method evaluates the ability of the model to generalize to new molecular scaffolds, providing a rigorous assessment of model performance on unseen chemical structures. However, for multi-task classification and regression datasets, we used random splitting. This approach was chosen due to the smaller sizes of these datasets, where scaffold splitting could result in imbalanced subsets or insufficient data for training and evaluation. Random splitting ensures that sufficient data is available in each subset while maintaining consistency across tasks.

Each dataset was split into training, validation, and test sets in an 8:1:1 ratio. The model was trained on the training set, with performance evaluated on the validation set after each training epoch. The model with the best validation performance was saved and used to compute results on the test set. Each experiment was repeated three times per dataset, with the mean and standard deviation of the results recorded. Detailed information about the datasets is provided in supplementary Table \ref{tab:datasetdetail}, while the hyperparameters used in the experiments are listed in supplementary Table \ref{tab:hyperparameter}.

For classification tasks, Area Under the Receiver Operating Characteristic curve (AUROC) and Area Under the Precision-Recall Curve (AUPRC) were used as evaluation metrics. For regression tasks, Root Mean Squared Error (RMSE) and Pearson correlation coefficient (PCC) were reported.

\subsection{Baselines}
We compare our proposed method against five baseline models: Directed Message Passing Neural Network (DMPNN), Fingerprints and Graph Neural Networks (FPGNN), Hierarchical Informative Graph Neural Networks (HiGNN), Deeper Graph Convolutional Network (DeeperGCN), and a transformer-based framework with focused attention for molecular representation (TransFoxMol). Each baseline represents a distinct approach to molecular representation learning.

\begin{itemize}
    \item \textbf{FPGNN}\citep{fpgnn2022}: Combines molecular fingerprints with features derived from graph attention networks, capturing both traditional cheminformatics features and structural insights from graphs.
    \item \textbf{DeeperGCN}\citep{deepergcn2020}: A pure graph neural network based on GCN, designed for deeper architectures to enhance feature extraction.
    \item \textbf{DMPNN}\citep{2019DMPNN}: Optimizes message passing by centering aggregation on bonds instead of atoms, effectively encoding the chemical structure and avoiding redundant loops.
    \item \textbf{HiGNN}\citep{HiGNN2022}: Using graph neural network to learn molecular representations at both the atomic level and the level of substructures. 
    \item \textbf{TransFoxMol}\citep{2023transfoxmol}: Integrates the power of graph neural networks and transformers to capture global and local molecular features efficiently.
\end{itemize}

\begin{table}[]
\captionsetup{justification=raggedright, singlelinecheck=false, font=large}
\caption{Performance evaluation on classification datasets.}
\label{tab:classification performance}
\centering
\setlength{\tabcolsep}{3pt}
\resizebox{\textwidth}{!}{
\begin{minipage}{\textwidth}
\raggedright
\begin{tabular}{|>{\centering\arraybackslash}p{2cm}|
                 >{\centering\arraybackslash}p{1cm}
                 >{\centering\arraybackslash}p{1.5cm}|cc|cc|cc|cc|cc|}
\hline
          & \multicolumn{2}{c|}{Sider}         & \multicolumn{2}{c|}{Tox21}         & \multicolumn{2}{c|}{Clintox}       & \multicolumn{2}{c|}{BBBP}          & \multicolumn{2}{c|}{BACE}          & \multicolumn{2}{c|}{HIV}           \\ \hline
          & \multicolumn{1}{c|}{AUROC} & {AUPRC} & \multicolumn{1}{c|}{AUROC} & {AUPRC} & \multicolumn{1}{c|}{AUROC} & {AUPRC} & \multicolumn{1}{c|}{AUROC} & {AUPRC} & \multicolumn{1}{c|}{AUROC} & {AUPRC} & \multicolumn{1}{c|}{AUROC} & {AUPRC} \\ \hline
FP-GNN    & \multicolumn{1}{c|}{\makecell{$0.661$ \\ $\pm 0.014$}} & {\makecell{$0.679$ \\ $\pm 0.026$}} & 
            \multicolumn{1}{c|}{\makecell{$0.833$ \\ $\pm 0.004$}} & {\makecell{$0.459$ \\ $\pm 0.018$}} & \multicolumn{1}{c|}{\makecell{$0.732$ \\ $\pm 0.068$}} & {\makecell{$0.622$ \\ $\pm 0.028$}} & \multicolumn{1}{c|}{\makecell{$0.892$ \\ $\pm 0.019$}} & {\makecell{$0.953$ \\ $\pm 0.007$}} &  
            \multicolumn{1}{c|}{\makecell{$0.852$ \\ $\pm 0.035$}} & {\makecell{$0.740$ \\ $\pm 0.042$}} & \multicolumn{1}{c|}{\makecell{$0.767$ \\ $\pm 0.039$}} & {\makecell{$0.328$ \\ $\pm 0.078$}} \\ \hline
DeeperGCN & \multicolumn{1}{c|}{\makecell{$0.622$ \\ $\pm 0.031$}} & {\makecell{$0.660$ \\ $\pm 0.025$}} & 
            \multicolumn{1}{c|}{\makecell{$0.840$ \\ $\pm 0.010$}} & {\makecell{$0.434$ \\ $\pm 0.021$}} & \multicolumn{1}{c|}{\makecell{$0.892$ \\ $\pm 0.048$}} & {\makecell{$\bm{0.741}$ \\ $\bm{\pm 0.048}$}} & \multicolumn{1}{c|}{\makecell{$0.860$ \\ $\pm 0.014$}} & {\makecell{$0.937$ \\ $\pm 0.008$}} &  
            \multicolumn{1}{c|}{\makecell{$0.830$ \\ $\pm 0.033$}} & {\makecell{$0.719$ \\ $\pm 0.039$}} & \multicolumn{1}{c|}{\makecell{$0.769$ \\ $\pm 0.041$}} & {\makecell{$0.300$ \\ $\pm 0.064$}} \\ \hline
DMPNN     & \multicolumn{1}{c|}{\makecell{$0.658$ \\ $\pm 0.032$}} & {\makecell{$0.680$ \\ $\pm 0.030$}} & 
            \multicolumn{1}{c|}{\makecell{$0.849$ \\ $\pm 0.006$}} & {\makecell{$0.481$ \\ $\pm 0.026$}} & \multicolumn{1}{c|}{\makecell{$0.895$ \\ $\pm 0.010$}} & {\makecell{$0.727$ \\ $\pm 0.062$}} & \multicolumn{1}{c|}{\makecell{$0.896$ \\ $\pm 0.014$}} & {\makecell{$0.956$ \\ $\pm 0.016$}} &  
            \multicolumn{1}{c|}{\makecell{$0.851$ \\ $\pm 0.028$}} & {\makecell{$0.742$ \\ $\pm 0.043$}} & \multicolumn{1}{c|}{\makecell{$0.758$ \\ $\pm 0.029$}} & {\makecell{$0.278$ \\ $\pm 0.043$}} \\ \hline
HiGNN     & \multicolumn{1}{c|}{\makecell{$0.656$ \\ $\pm 0.024$}} & {\makecell{$0.669$ \\ $\pm 0.027$}} & 
            \multicolumn{1}{c|}{\makecell{$0.844$ \\ $\pm 0.006$}} & {\makecell{$0.462$ \\ $\pm 0.018$}} & \multicolumn{1}{c|}{\makecell{$0.889$ \\ $\pm 0.026$}} & {\makecell{$0.735$ \\ $\pm 0.070$}} & \multicolumn{1}{c|}{\makecell{$0.892$ \\ $\pm 0.014$}} & {\makecell{$0.943$ \\ $\pm 0.017$}} &  
            \multicolumn{1}{c|}{\makecell{$0.836$ \\ $\pm 0.029$}} & {\makecell{$0.740$ \\ $\pm 0.048$}} & \multicolumn{1}{c|}{\makecell{$0.768$ \\ $\pm 0.038$}} & {\makecell{$0.310$ \\ $\pm 0.066$}} \\ 
            \hline
            
TransFoxMol      & \multicolumn{1}{c|}{\makecell{$0.636$ \\ $\pm 0.022$}} & {\makecell{$0.686$ \\ $\pm 0.040$}} & 
            \multicolumn{1}{c|}{\makecell{$0.816$ \\ $\pm 0.011$}} & {\makecell{$0.367$ \\ $\pm 0.011$}} & \multicolumn{1}{c|}{\makecell{$0.830$ \\ $\pm 0.047$}} & {\makecell{$0.624$ \\ $\pm 0.036$}} & \multicolumn{1}{c|}{\makecell{$0.881$ \\ $\pm 0.015$}} & {\makecell{$0.947$ \\ $\pm 0.005$}} &  
            \multicolumn{1}{c|}{\makecell{$0.801$ \\ $\pm 0.054$}} & {\makecell{$0.693$ \\ $\pm 0.079$}} & \multicolumn{1}{c|}{\makecell{$0.727$ \\ $\pm 0.037$}} & {\makecell{$0.232$ \\ $\pm 0.063$}} \\ 
            \hline

\makecell{MolGraph-\\xLSTM (Ours)}  & 
\multicolumn{1}{c|}{\makecell{$\bm{0.697}$ \\ $\bm{\pm 0.022}$}} & {\makecell{$\bm{0.713}$ \\ $\bm{\pm 0.032}$}} & \multicolumn{1}{c|}{\makecell{$\bm{0.854}$ \\ $\bm{\pm 0.003}$}} & {\makecell{$\bm{0.487}$ \\ $\bm{\pm 0.045}$}} & \multicolumn{1}{c|}{\makecell{$\bm{0.904}$ \\ $\bm{\pm 0.032}$}} & {\makecell{$0.714$ \\ $\pm 0.026$}} & 
\multicolumn{1}{c|}{\makecell{$\bm{0.959}$ \\ $\bm{\pm 0.006}$}} & {\makecell{$\bm{0.987}$ \\ $\bm{\pm 0.002}$}} & \multicolumn{1}{c|}{\makecell{$\bm{0.869}$ \\ $\bm{\pm 0.016}$}} & {\makecell{$\bm{0.784}$ \\ $\bm{\pm 0.029}$}} & 
\multicolumn{1}{c|}{\makecell{$\bm{0.775}$ \\ $\bm{\pm 0.027}$}} & {\makecell{$\bm{0.355}$ \\ $\bm{\pm 0.050}$}} \\ 
\hline

\end{tabular}
\end{minipage}
}
\end{table}

\begin{table}[]
\centering
\setlength{\tabcolsep}{3pt}
\begin{minipage}{\textwidth}
\captionsetup{justification=raggedright, singlelinecheck=false, font=normalsize}
\caption{Performance evaluation on regression datasets.}
\label{tab:regressionperformance}
\begin{tabular}{|c|cc|cc|cc|cc|}
\hline
                       & \multicolumn{2}{c|}{ESOL}                         & \multicolumn{2}{c|}{Lipo}                         & \multicolumn{2}{c|}{Freesolv}                     &
                       \multicolumn{2}{c|}{Caco2}
                       \\ \hline
                       & \multicolumn{1}{c|}{RMSE} & {PCC}                    & \multicolumn{1}{c|}{RMSE} & {PCC}                    & \multicolumn{1}{c|}{RMSE} & {PCC}                    &
                       \multicolumn{1}{c|}{RMSE} & {PCC} 
                       \\ \hline
FP-GNN                 
& \multicolumn{1}{c|}{\makecell{$0.658$ \\ $\pm 0.006$}} & {\makecell{$0.946$ \\ $\pm 0.006$}} &     
\multicolumn{1}{c|}{{\makecell{$0.610$ \\ $\pm 0.028$}}} & {\makecell{$0.861$ \\ $\pm 0.012$}} & 
\multicolumn{1}{c|}{{\makecell{$1.106$ \\ $\pm 0.195$}}} & {\makecell{$0.951$ \\ $\pm 0.023$}} &
\multicolumn{1}{c|}{{\makecell{$\bm{0.491}$ \\ $\bm{\pm 0.023}$}}} & {\makecell{$\bm{0.785}$ \\ $\bm{\pm 0.015}$}} 
\\ \hline
DeeperGCN              
& \multicolumn{1}{c|}{\makecell{$0.615$ \\ $\pm 0.044$}} & {\makecell{$0.954$ \\ $\pm 0.008$}} & 
\multicolumn{1}{c|}{{\makecell{$0.645$ \\ $\pm 0.048$}}} & {\makecell{$0.842$ \\ $\pm 0.026$}} & 
\multicolumn{1}{c|}{{\makecell{$1.261$ \\ $\pm 0.022$}}} & {\makecell{$0.938$ \\ $\pm 0.007$}} &
\multicolumn{1}{c|}{{\makecell{$0.521$ \\ $\pm 0.013$}}} & {\makecell{$0.752$ \\ $\pm 0.014$}} 
\\ \hline
DMPNN                  
& \multicolumn{1}{c|}{\makecell{$0.575$ \\ $\pm 0.073$}} & {\makecell{$0.957$ \\ $\pm 0.015$}} & 
\multicolumn{1}{c|}{\makecell{$0.553$ \\ $\pm 0.033$}}  & {\makecell{$0.842$ \\ $\pm 0.026$}} & 
\multicolumn{1}{c|}{\makecell{$1.211$ \\ $\pm 0.120$}}  & {\makecell{$0.945$ \\ $\pm 0.007$}} &
\multicolumn{1}{c|}{\makecell{$0.530$ \\ $\pm 0.020$}}  & {\makecell{$0.746$ \\ $\pm 0.020$}} 
\\ \hline
HiGNN                  
& \multicolumn{1}{c|}{\makecell{$0.570$ \\ $\pm 0.061$}} & {\makecell{$0.959$ \\ $\pm 0.013$}} &
\multicolumn{1}{c|}{\makecell{$0.563$ \\ $\pm 0.041$}} & {\makecell{$0.882$ \\ $\pm 0.018$}} & 
\multicolumn{1}{c|}{\makecell{$1.068$ \\ $\pm 0.092$}} & {\makecell{$0.956$ \\ $\pm 0.007$}} &
\multicolumn{1}{c|}{\makecell{$0.507$ \\ $\pm 0.010$}} & {\makecell{$0.771$ \\ $\pm 0.007$}}
\\ \hline
TransFoxMol                  
& \multicolumn{1}{c|}{\makecell{$0.930$ \\ $\pm 0.261$}} & {\makecell{$0.917$ \\ $\pm 0.047$}} &
\multicolumn{1}{c|}{\makecell{$0.652$ \\ $\pm 0.033$}} & {\makecell{$0.855$ \\ $\pm 0.011$}} & 
\multicolumn{1}{c|}{\makecell{$1.225$ \\ $\pm 0.155$}} & {\makecell{$0.945$ \\ $\pm 0.007$}} &
\multicolumn{1}{c|}{\makecell{$0.545$ \\ $\pm 0.026$}} & {\makecell{$0.735$ \\ $\pm 0.025$}}
\\ \hline
\makecell{MolGraph-\\ xLSTM (Ours)}   & 
\multicolumn{1}{c|}{\makecell{$\bm{0.527}$ \\ $\bm{\pm 0.046}$}} & {\makecell{$\bm{0.965}$ \\ $\bm{\pm 0.010}$}} &
\multicolumn{1}{c|}{\makecell{$\bm{0.550}$ \\ $\bm{\pm 0.026}$}} & {\makecell{$\bm{0.888}$ \\ $\bm{\pm 0.011}$}} & 
\multicolumn{1}{c|}{\makecell{$\bm{1.024}$ \\ $\bm{\pm 0.076}$}} & {\makecell{$\bm{0.960}$ \\ $\bm{\pm 0.006}$}} &
\multicolumn{1}{c|}{\makecell{$0.503$ \\ $\pm 0.004$}} & {\makecell{$0.771$ \\ $\pm 0.010$}}
\\ \hline
\end{tabular}
\end{minipage}
\end{table}

\subsection{Experimental results}
MolGraph-xLSTM demonstrates improved performance across both classification and regression datasets, highlighting its robustness in handling diverse molecular property prediction tasks. In the classification tasks (Table \ref{tab:classification performance}), MolGraph-xLSTM achieves particularly strong results on the Sider and BBBP datasets. For the Sider dataset, MolGraph-xLSTM achieves an AUROC of 0.697 $\pm$ 0.022, representing a 5.45\% improvement over the best baseline, FP-GNN (0.661 $\pm$ 0.014). Similarly, on the BBBP dataset, MolGraph-xLSTM achieves an AUROC of 0.959 $\pm$ 0.006, which is a 7.03\% improvement compared to the best baseline, TransFoxMol (0.896 $\pm$ 0.024). 

For regression datasets (Table \ref{tab:regressionperformance}), MolGraph-xLSTM delivers competitive performance across multiple benchmarks. On the ESOL dataset, MolGraph-xLSTM achieves an RMSE of 0.527 $\pm$ 0.046, reflecting a 7.54\% improvement over the best-performing baseline, HiGNN (0.570 $\pm$ 0.061). On the FreeSolv dataset, MolGraph-xLSTM achieves the lowest RMSE of 1.024 $\pm$ 0.076 and the highest PCC of 0.960 $\pm$ 0.006, demonstrating its reliability in regression tasks.

\begin{figure}[htbp] 
  \centering
  \begin{subfigure}[b]{\textwidth}
    \centering
    \includegraphics[width=\textwidth]{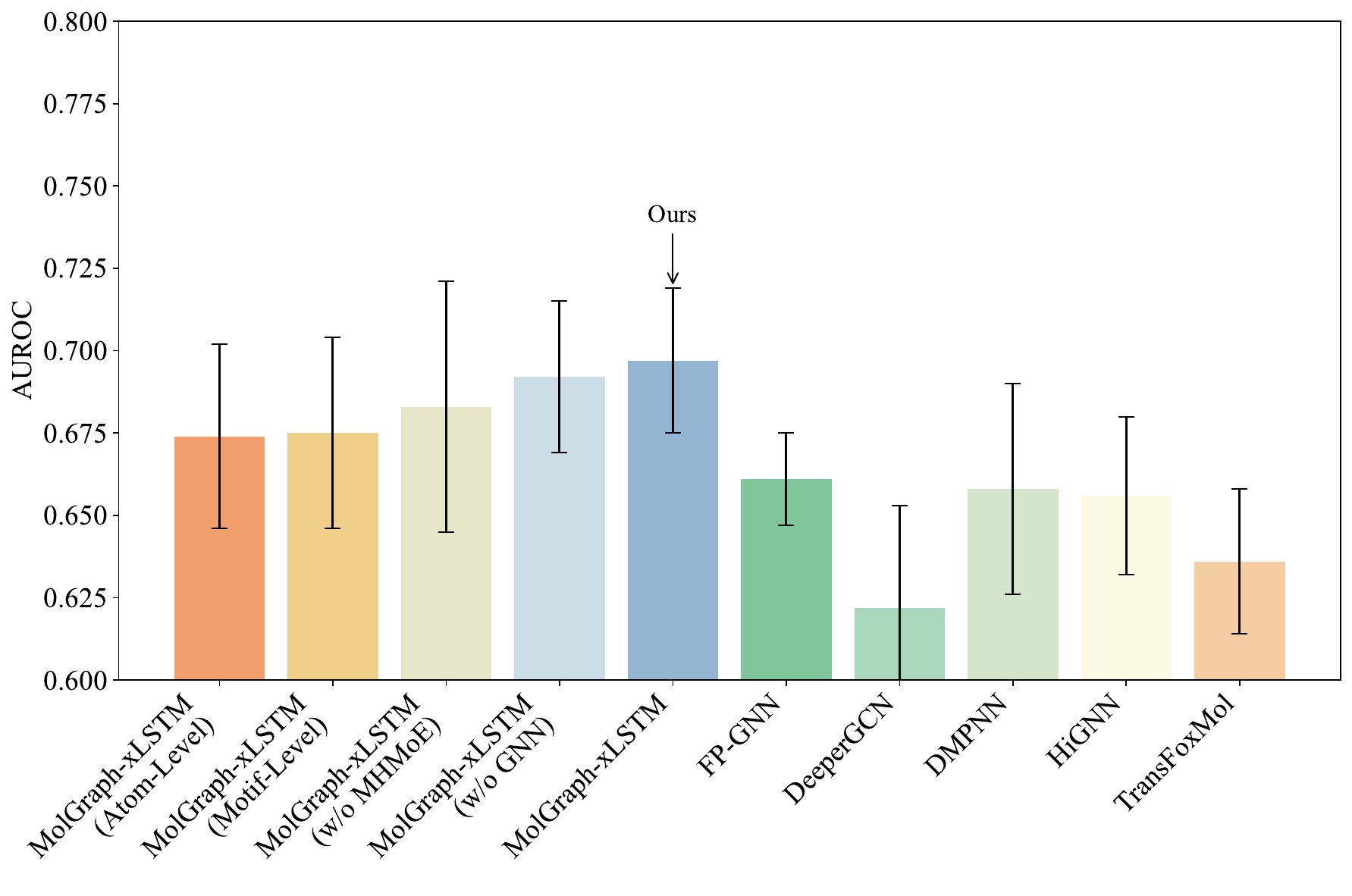}
    \caption{Sider dataset}
  \end{subfigure}
  \hfill
  \begin{subfigure}[b]{\textwidth}
    \centering
    \includegraphics[width=\textwidth]{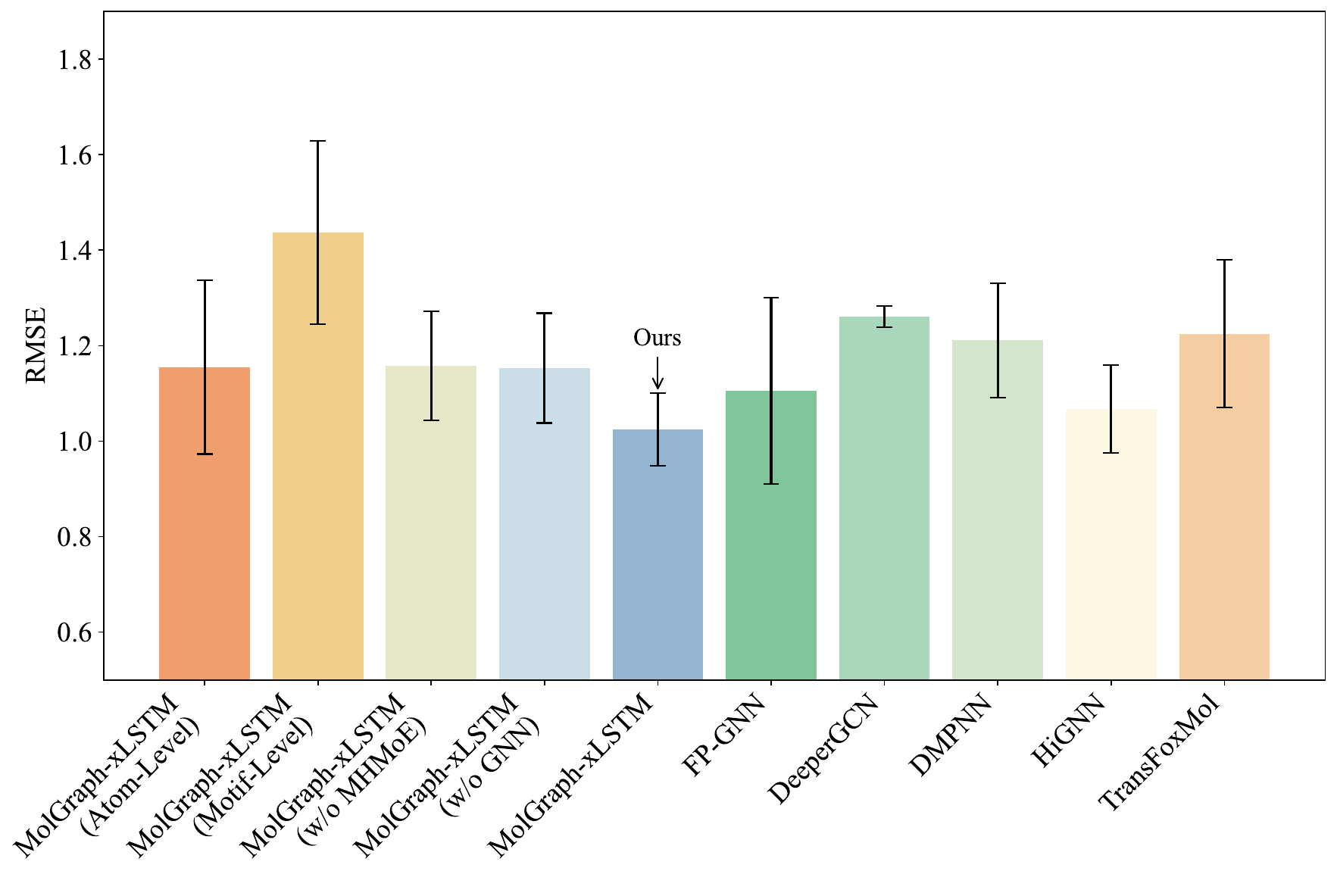}
    \caption{Freesolv dataset}
  \end{subfigure}
  \caption{\textbf{Ablation study results on the Sider and FreeSolv datasets.} Performance comparison of ablation variants against the full MolGraph-xLSTM model and baseline models.}
  \label{figure:ablation}
\end{figure}

\begin{table}[]
\centering
\caption{Ablation results for MolGraph-xLSTM on the Sider(classification) and FreeSolv(regression) datasets.}
\label{table:ablation}
\begin{minipage}{\textwidth}
\raggedright
\begin{tabular}{|>{\centering\arraybackslash}m{2.5cm}
                 |>{\centering\arraybackslash}m{1cm}
                 >{\centering\arraybackslash}m{1cm}|>{\centering\arraybackslash}m{1cm}|>{\centering\arraybackslash}m{1cm}|}
\hline
          & \multicolumn{2}{c|}{Sider}           
          & \multicolumn{2}{c|}{Freesolv} \\ \hline
          & \multicolumn{1}{c|}{AUROC} & {AUPRC} 
          & \multicolumn{1}{c|}{RMSE} & {PCC} \\ \hline
MolGraph-xLSTM (Atom-Level)  & 
\multicolumn{1}{c|}{\makecell{$0.674$ \\ $\pm 0.028$}} & {\makecell{$0.697$ \\ $\pm 0.034$}} & 
\multicolumn{1}{c|}{\makecell{$1.155$ \\ $\pm 0.182$}} & {\makecell{$0.947$ \\ $\pm 0.017$}}\\ 
\hline

MolGraph-xLSTM (Motif-Level)  & 
\multicolumn{1}{c|}{\makecell{$0.675$ \\ $\pm 0.029$}} & {\makecell{$0.699$ \\ $\pm 0.030$}} & 
\multicolumn{1}{c|}{\makecell{$1.437$ \\ $\pm 0.192$}} & {\makecell{$0.924$ \\ $\pm 0.015$}}
\\ 
\hline

MolGraph-xLSTM (w/o MHMoE) &
\multicolumn{1}{c|}{\makecell{$0.683$ \\ $\pm 0.038$}} & {\makecell{$0.704$ \\ $\pm 0.034$}} & 
\multicolumn{1}{c|}{\makecell{$1.158$ \\ $\pm 0.114$}} & {\makecell{$0.949$ \\ $\pm 0.010$}}
\\
\hline

MolGraph-xLSTM (w/o GNN) &
\multicolumn{1}{c|}{\makecell{$0.692$ \\ $\pm 0.023$}} & {\makecell{$0.708$ \\ $\pm 0.031$}} & 
\multicolumn{1}{c|}{\makecell{$1.153$ \\ $\pm 0.115$}} & {\makecell{$0.949$ \\ $\pm 0.014$}}
\\
\hline

\end{tabular}
\end{minipage}

\end{table}

\subsection{Ablation Study}
\subsubsection{Effect of Different Designed Modules} \label{sec4.4.1 ablation1}
We conducted an ablation study to evaluate the contributions of different components in MolGraph-xLSTM, including the atom-level branch (MolGraph-xLSTM (Atom-Level)), motif-level branch (MolGraph-xLSTM (Motif-LeveL)), multi-head mixture-of-experts module  (MolGraph-xLSTM(w/o MHMoE)), and the GNN component within the atom-level branch  (MolGraph-xLSTM(w/o GNN)). The results, presented in Table~\ref{table:ablation} and Figure \ref{figure:ablation}, highlight the importance of these components in achieving superior performance.

The full MolGraph-xLSTM model consistently outperformed all ablation variants, highlighting the effectiveness of its integrated architecture. Notably, even with only the atom-level branch, MolGraph-xLSTM achieved competitive performance, outperforming other atom-level graph-based models like DMPNN and DeeperGCN, as well as TransFoxMol, a hybrid model integrating GNN and Transformer. These results validate the design of our hybrid GNN and xLSTM framework as an effective approach for molecular representation learning. For the motif-level branch, it also outperformed other baselines on the Sider dataset, including HiGNN, which also utilizes motif-level graphs, in the classification task. However, its performance on the regression dataset was suboptimal. This suggests that the motif-level initialization features utilized in our model may not sufficiently capture the granularity required for regression tasks, highlighting opportunities for further improvement.

The MHMoE module contributed to the model performance, particularly on the FreeSolv dataset. Removing the MHMoE module resulted in an RMSE increase from 1.024 to 1.158, closely aligning with the performance of the atom-level-only variant, indicating its role in improving regression performance. As shown in Figure~\ref{fig:freesolv_moe_activation} and Figure~\ref{fig:sider_moe_activation}, the activation maps demonstrate that all experts actively contribute to the task, indicating effective load balancing. This balanced activation ensures no single expert is overwhelmed, allowing the network to fully leverage the diverse expertise of all experts. 

Among the four components, the GNN had the least impact on the Sider dataset but showed a notable influence on FreeSolv. Overall, the ablation study demonstrates that the atom- and motif-level branches provide complementary insights into molecular representation learning, and their integration enhances the model performance. This highlights the effectiveness of the proposed approach for molecular modeling.

\subsubsection{Impact of Node Input Order for Molecular Graphs on Performance}
\begin{figure}[htbp]
  \centering
  \begin{subfigure}[b]{0.45\textwidth}
    \centering
    \includegraphics[width=\textwidth]{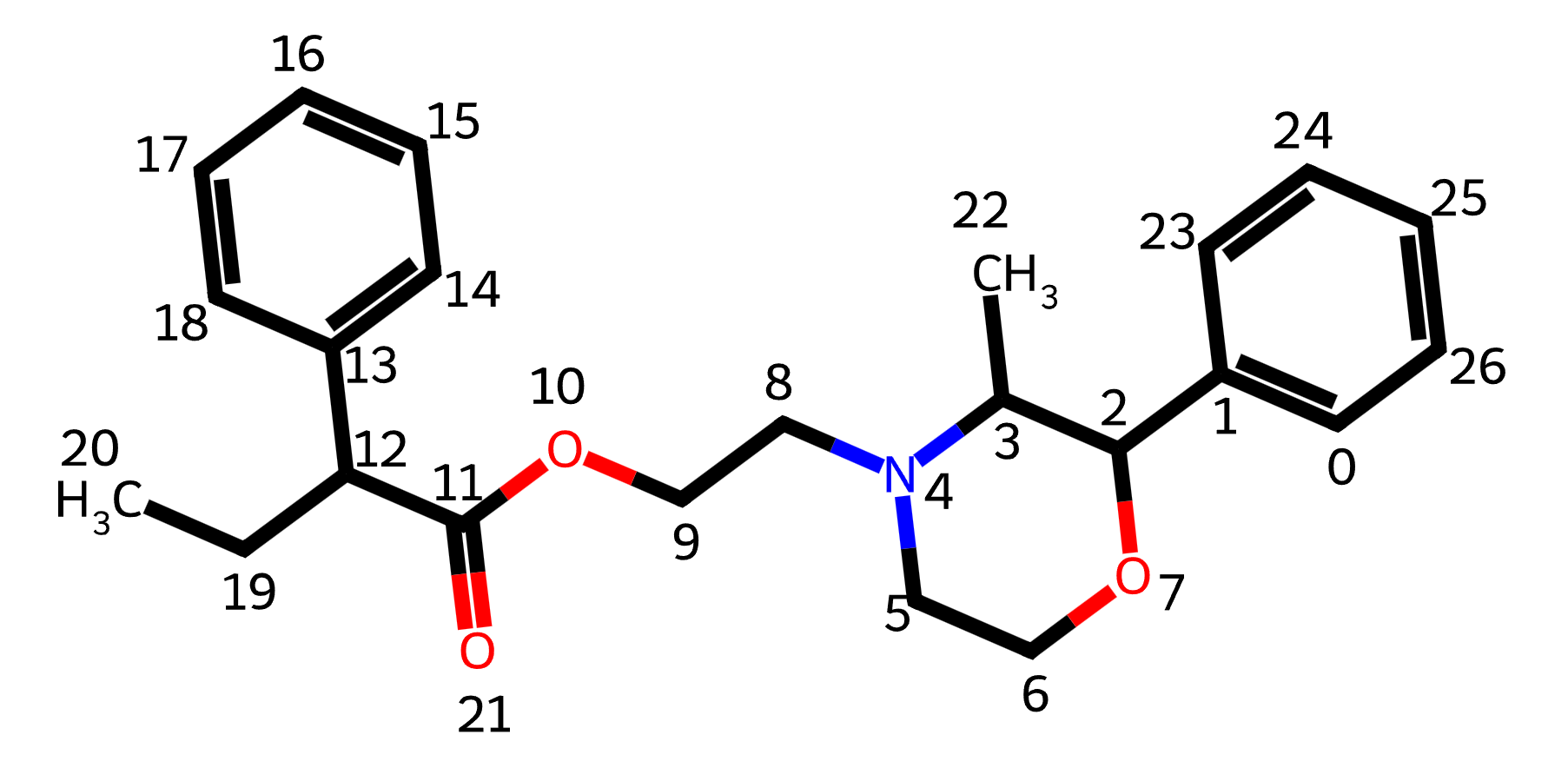}
    \caption{RDKit Default Order}
  \end{subfigure}
  \hfill
  \begin{subfigure}[b]{0.45\textwidth}
    \centering
    \includegraphics[width=\textwidth]{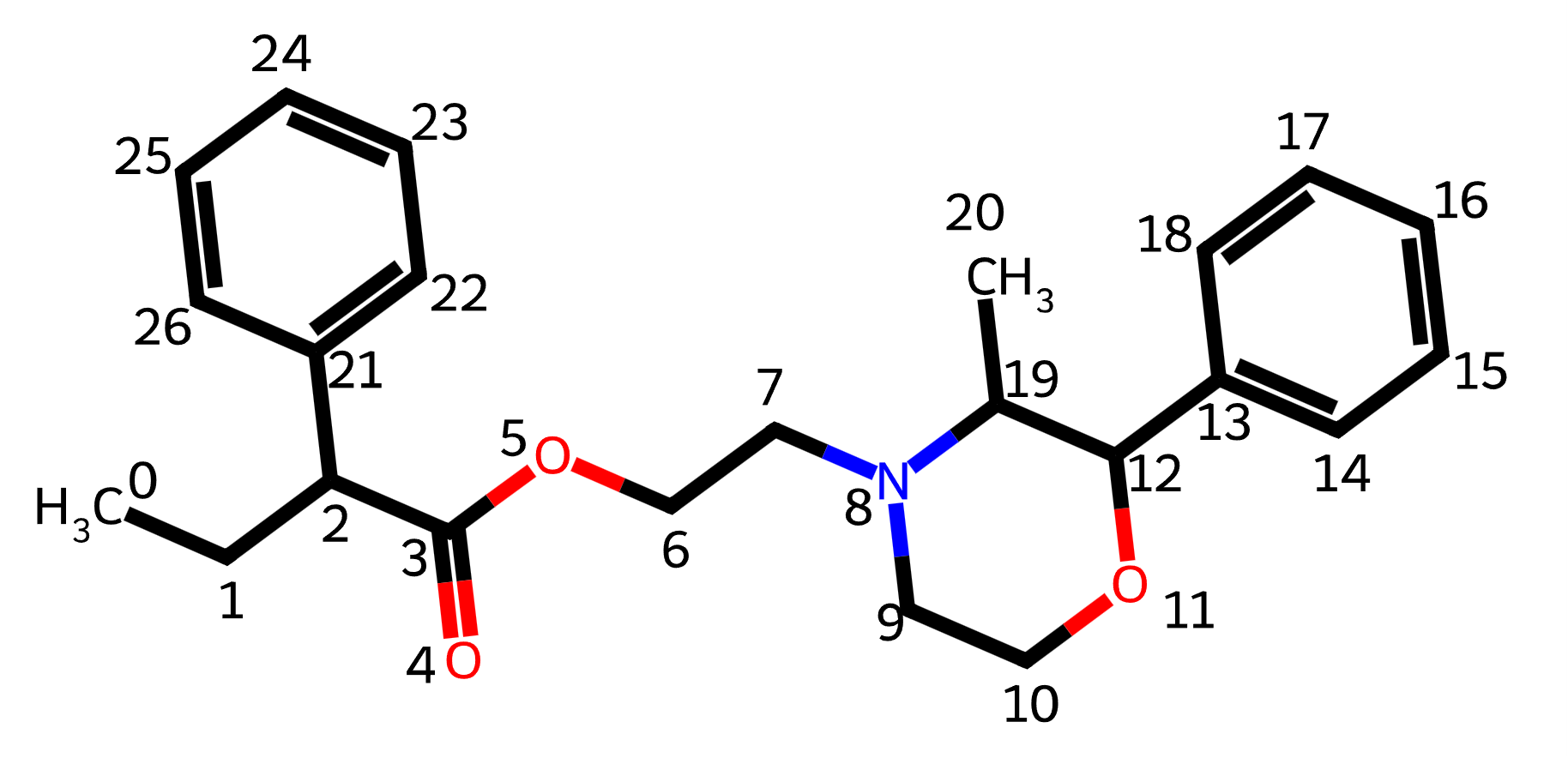}
    \caption{DFS order}
  \end{subfigure}
  \caption{\textbf{Examples of different atom input orders for molecular graphs in xLSTM.} (a) RDKit Default Order: Atoms are ordered as per the default output from RDKit. (b) DFS Order: Atoms are ordered based on a Depth-First Search traversal of the molecular graph.}
  \label{figure:differentorders}
\end{figure}

\begin{figure}[htbp]
  \centering
  \begin{subfigure}[b]{0.45\textwidth}
    \centering
    \includegraphics[width=\textwidth]{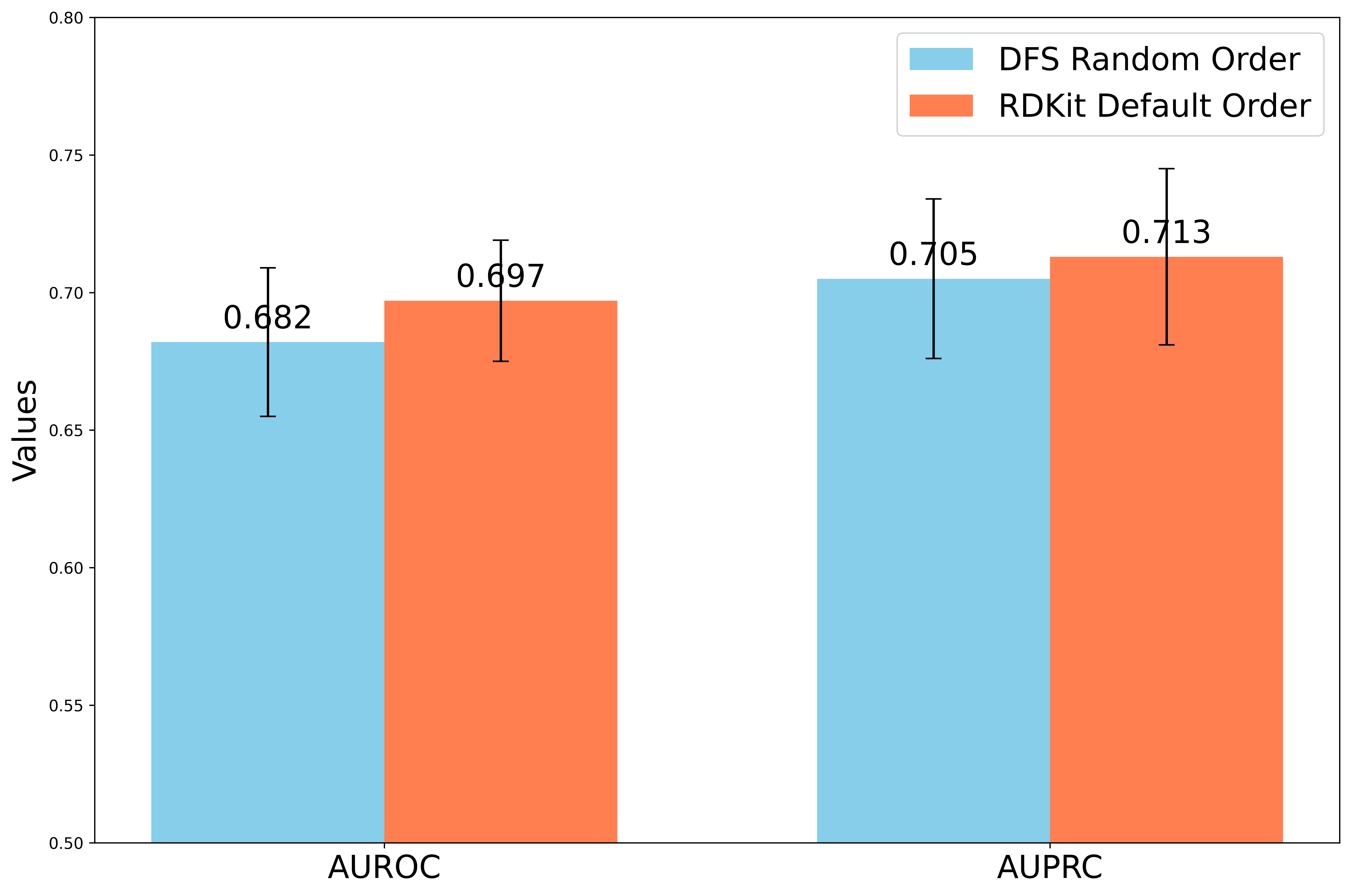}
    \caption{Sider dataset}
  \end{subfigure}
  \hfill
  \begin{subfigure}[b]{0.45\textwidth}
    \centering
    \includegraphics[width=\textwidth]{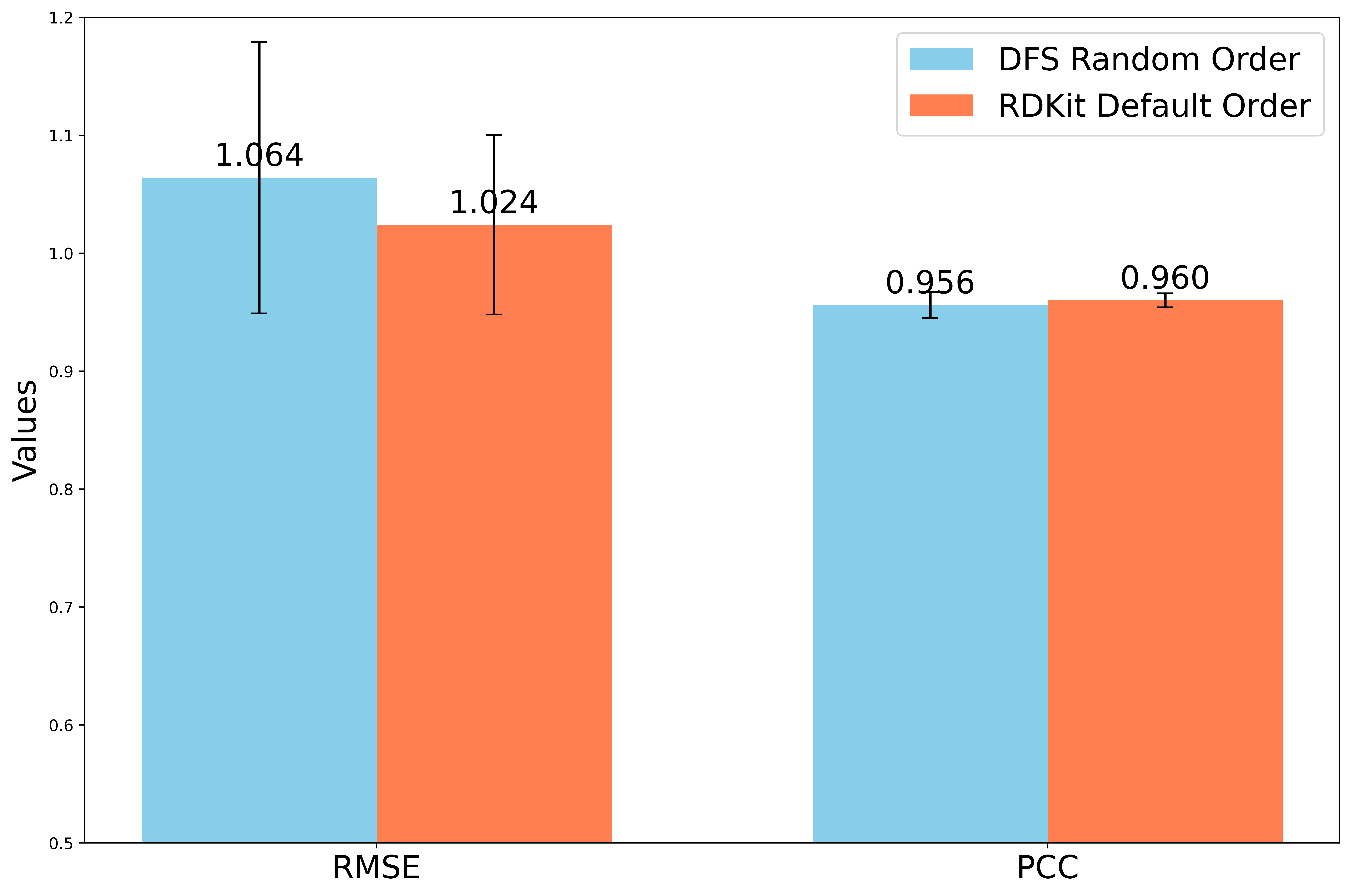}
    \caption{Freesolv dataset}
  \end{subfigure}
  \caption{\textbf{Performance comparison of MolGraph-xLSTM trained with different node orderings.} The RDKit Default Order refers to the node sequence provided by RDKit, while the DFS Random Order generates a sequence by performing a depth-first search starting from a randomly selected node. (a) Results on the Sider dataset (classification) using AUROC and AUPRC as metrics. (b) Results on the FreeSolv dataset (regression) using RMSE and PCC as metrics. }
  \label{figure:order}
\end{figure}

xLSTM is originally designed for sequence data, which inherently has a fixed order. However, graph data does not have this property, as it can start from any node (Figure \ref{figure:differentorders}). In our initial tests, we used the default node order provided by RDKit. In this section, we evaluate the effect of using a randomized starting node during training. Specifically, we generate the node sequence by performing a depth-first search (DFS) starting from a randomly selected initial node in the graph for each training instance.

Figure \ref{figure:order} compares the performance of MolGraph-xLSTM trained with the RDKit default node order and the DFS random order on Sider and Freesolv datasets. On the Sider dataset (Figure \ref{figure:order} (a)), the model trained with the RDKit default order slightly outperformed the DFS random order in both AUROC and AUPRC metrics. Similarly, on the FreeSolv dataset (Figure \ref{figure:order} (b)), the RMSE and PCC metrics indicate a marginal advantage for the RDKit default order. Despite these differences, the results show that MolGraph-xLSTM achieves competitive performance with both node orderings. This suggests that the model is robust to changes in the input node sequence.

\begin{figure}[htbp]
\centering
\includegraphics[width=1\textwidth]{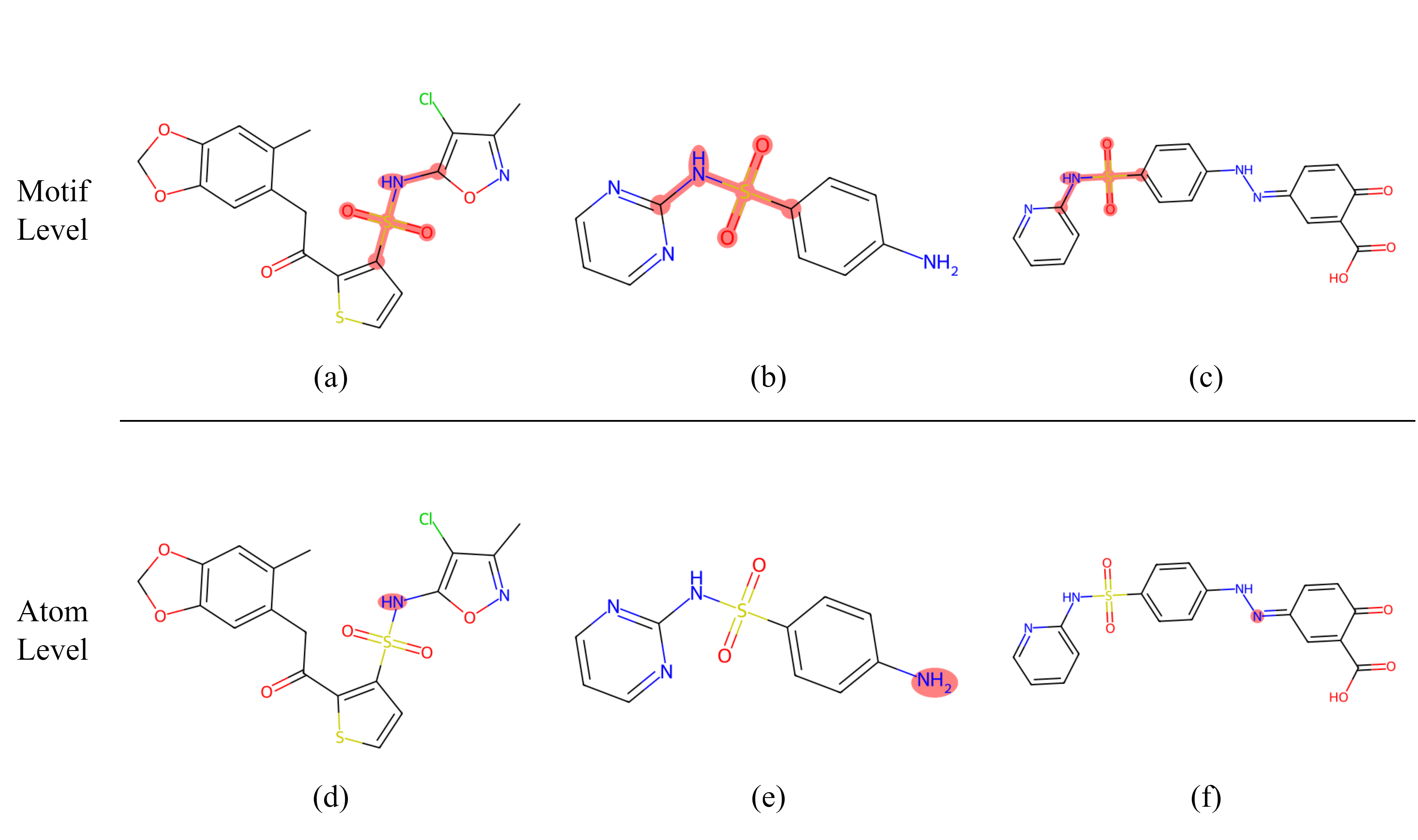}
\caption{\textbf{Visualization of the highest-weighted motifs and atoms identified by the model for molecules from the Sider test set containing the $\bm{SO_2NH}$ substructure.} The top row highlights the motifs with the highest attention weights from the motif-level branch, while the bottom row highlights the atoms with the highest attention weights from the atom-level branch.}
\label{figure:SO2NH}
\end{figure}

\begin{figure}[htbp]
\centering
\includegraphics[width=1\textwidth]{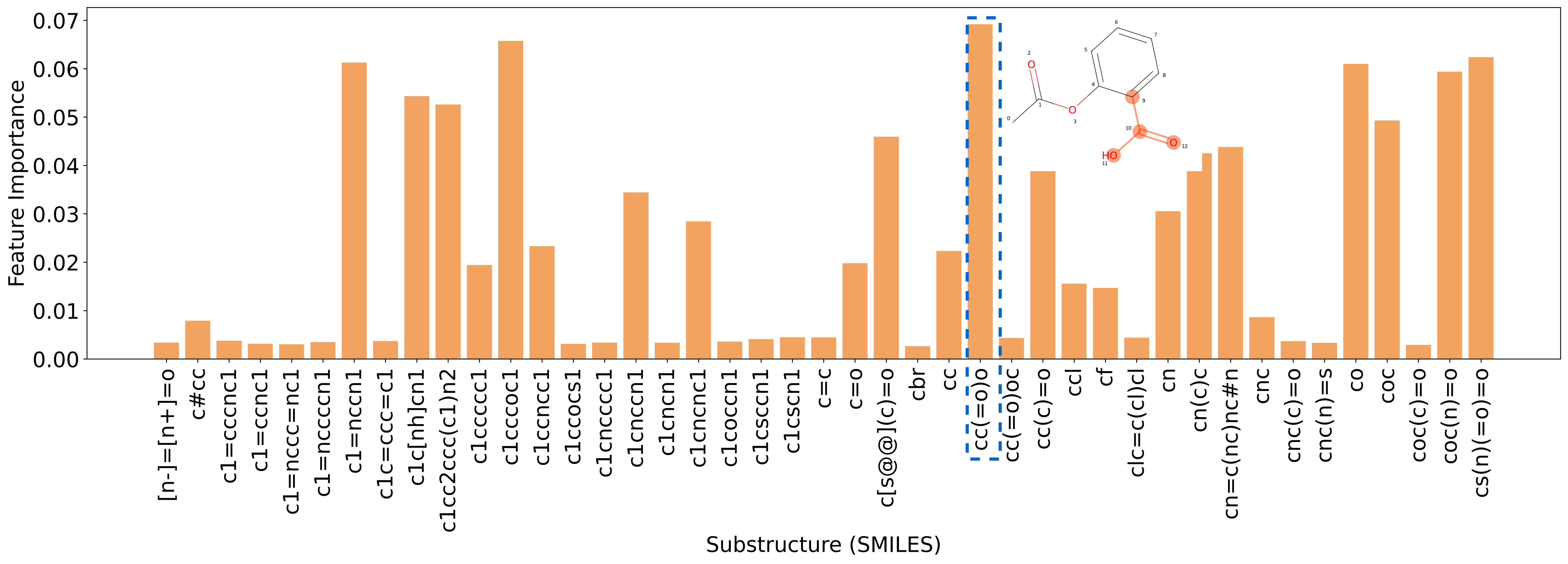}
\caption{\textbf{Importance scores of substructures identified by MolGraph-xLSTM on the BBBP dataset.} For each molecule, the substructure with the highest model-assigned weight was analyzed using a random forest model to determine its relationship with BBBP labels. The substructure $-CC(=O)O-$, containing a carboxylic group, received the highest importance score (highlighted by the blue dashed box).}
\label{figure:c(=o)o}
\end{figure}

\subsection{Interpretability analysis} \label{section: Interpretability}
To evaluate the interpretability of MolGraph-xLSTM, we visualized the motifs and atomic sites with the highest model-assigned weights from the motif-level and atom-level networks. By applying max-pooling to the output of the xLSTM layer, we identified the features with the greatest contributions, providing us insight into the substructures and atomic sites that are most closely related to the properties of a particular molecule.

In Figure \ref{figure:SO2NH}, all three molecules highlight the $-SO_2NH-$ (sulfonamide) substructure, a chemical motif known to be strongly linked with adverse reactions such as Type IV hypersensitivity, blurred vision, and other side effects \citep{2024sulfonamides}. These adverse effects correspond to side effects labeled in the Sider dataset, including Eye Disorders, Immune System Disorders, and Skin and Subcutaneous Tissue Disorders, demonstrating an alignment between the highlighted substructure and known biological properties of sulfonamides. Additionally, molecules like Figure \ref{figure:SO2NH}(e) and Figure \ref{figure:SO2NH}(f) emphasize atomic sites beyond the sulfonamide motif. In Figure \ref{figure:SO2NH}(f), the highlighted $N$ atom resides within the hydrazine group ($-NH-N=$), which is known to exert toxic effects on multiple organ systems, including neurological, hematological and pulmonary \citep{2023Hydrazine}. This suggests that the atom-level network captures additional fine-grained features that complement the broader motif-level representations, demonstrating the capacity of the model to integrate complementary information from both atom-level and motif-level networks.

We further conducted an analysis on the BBBP dataset (Blood-Brain Barrier Permeability), a crucial property in evaluating the ability of a drug to cross the blood-brain barrier and target central nervous system (CNS) disorders. Accurate prediction of this property is essential for developing CNS-targeted therapies. For each molecule in the dataset, the substructure with the highest weight assigned by MolGraph-xLSTM was identified. These substructures were further analyzed using a random forest model  \citep{2001RF} to determine their relationship with BBBP labels.

Figure \ref{figure:c(=o)o} illustrates the importance scores of substructures as determined by the random forest model. Among these, the substructure $-CC(=O)O-$, containing a carboxylic group ($-C(=O)O-$), achieved the highest importance score. This finding is supported by previous studies \citep{2016bbbp1, 2020bbbp2}, which have highlighted the role of the carboxylic group in influencing BBBP. 

\begin{figure}[htbp]
  \centering
  \begin{subfigure}[b]{0.45\textwidth}
    \centering
    \includegraphics[width=\textwidth]{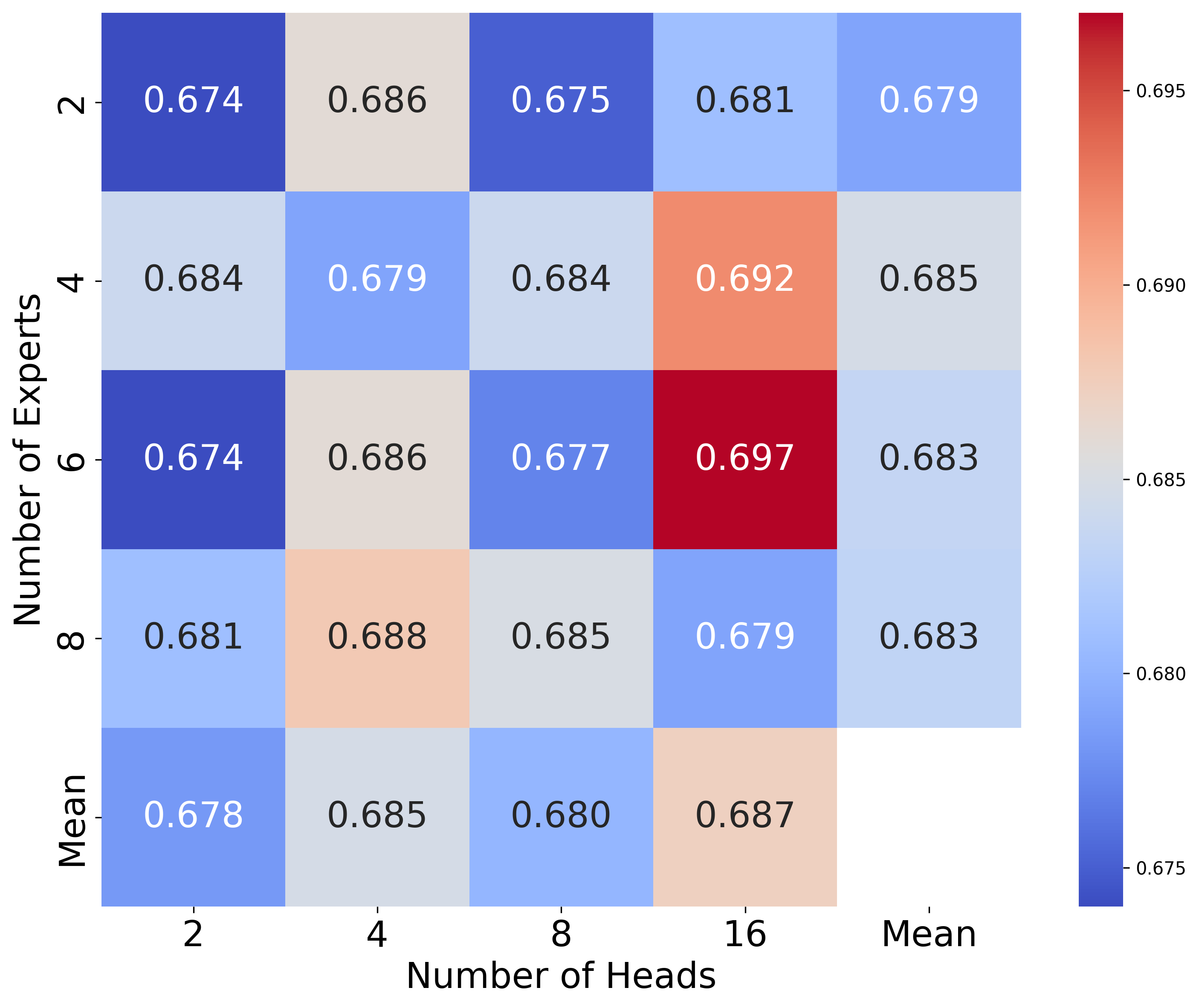}
    \caption{Sider dataset (AUROC)}
  \end{subfigure}
  \hfill
  \begin{subfigure}[b]{0.45\textwidth}
    \centering
    \includegraphics[width=\textwidth]{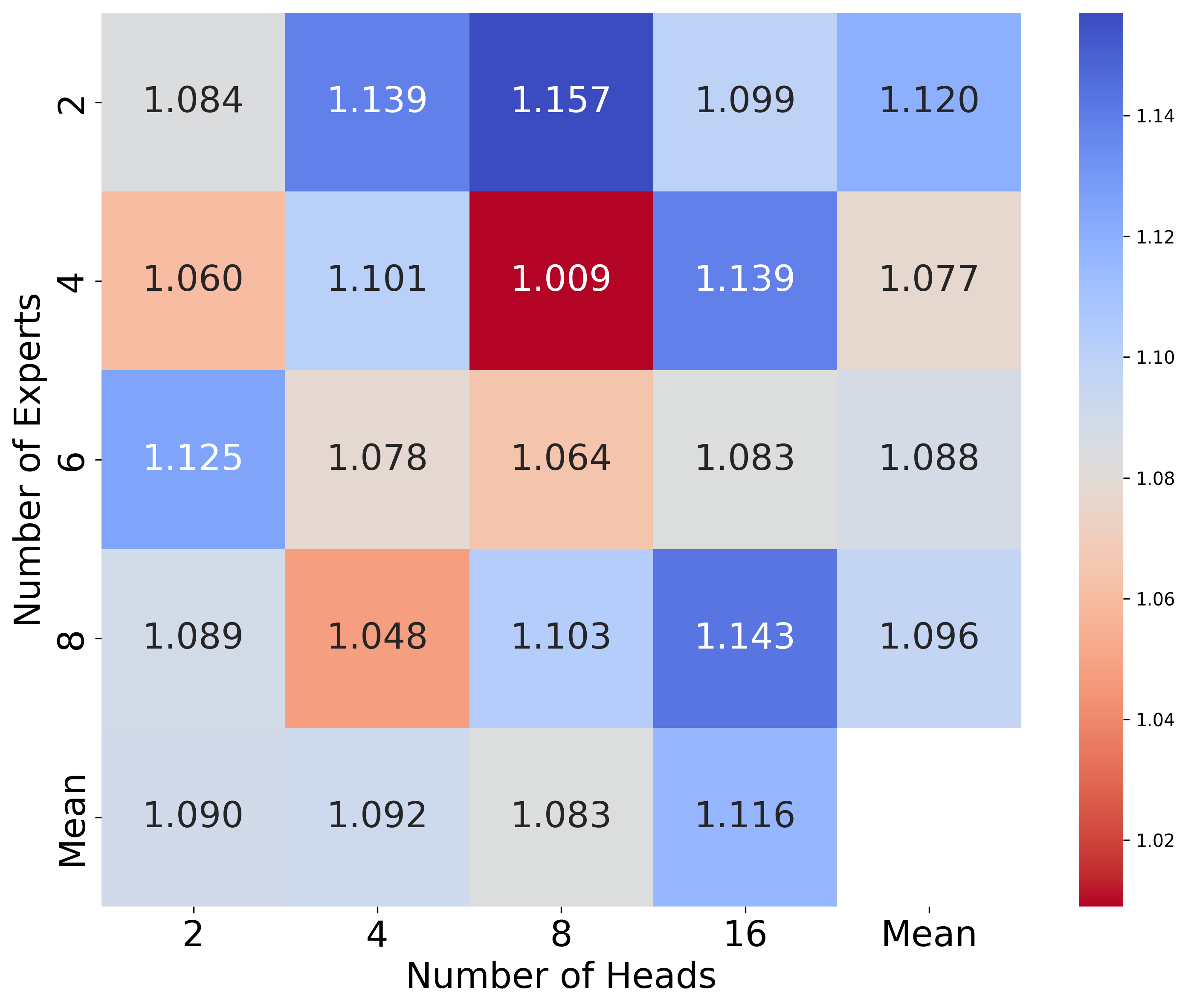}
    \caption{Freesolv dataset (RMSE)}
  \end{subfigure}
  \caption{\textbf{Analysis of the impact of the number of experts and heads on the Sider and FreeSolv datasets.} (a) Performance on the Sider dataset measured by AUROC, where red indicates higher AUROC (better performance) and blue represents lower AUROC (worse performance). (b) Performance on the FreeSolv dataset measured by RMSE, with red corresponding to lower RMSE (better performance) and blue indicating higher RMSE (worse performance).}
  \label{figure: headandexpert}
\end{figure}

\subsection{Hyperparameter analysis}
\subsubsection{Performance of MolGraph-xLSTM with Varing Number of Experts and Heads in the MHMoE}

The heatmaps in Figure \ref{figure: headandexpert} reveal the impact of the number of experts and heads in the MHMoE module on the model's performance for the Sider and FreeSolv datasets. For both datasets, configurations with 2 experts generally perform poorly, while increasing the number of experts to 4 or 6 yields better results. Beyond 6 experts, no significant improvements are observed, suggesting that additional experts may become redundant for these datasets, as they do not process substantially different information.

For the Sider dataset, measured by AUROC, an increase in the number of heads consistently enhances performance, indicating that more heads improve the model's ability to handle classification tasks. In contrast, for the FreeSolv dataset, measured by RMSE, increasing the number of heads beyond 8 leads to a noticeable decline in performance, particularly when the number of heads reaches 16. This decline is likely due to overfitting, as FreeSolv is a relatively small dataset. These observations highlight the need to balance the number of experts and heads based on the task and dataset size, as excessive complexity can negatively affect performance.

\subsubsection{Performance of MolGraph-xLSTM with Varing Number of Jump Layers}
\begin{figure}[htbp]
  \centering
  \begin{subfigure}[b]{0.45\textwidth}
    \centering
    \includegraphics[width=\textwidth]{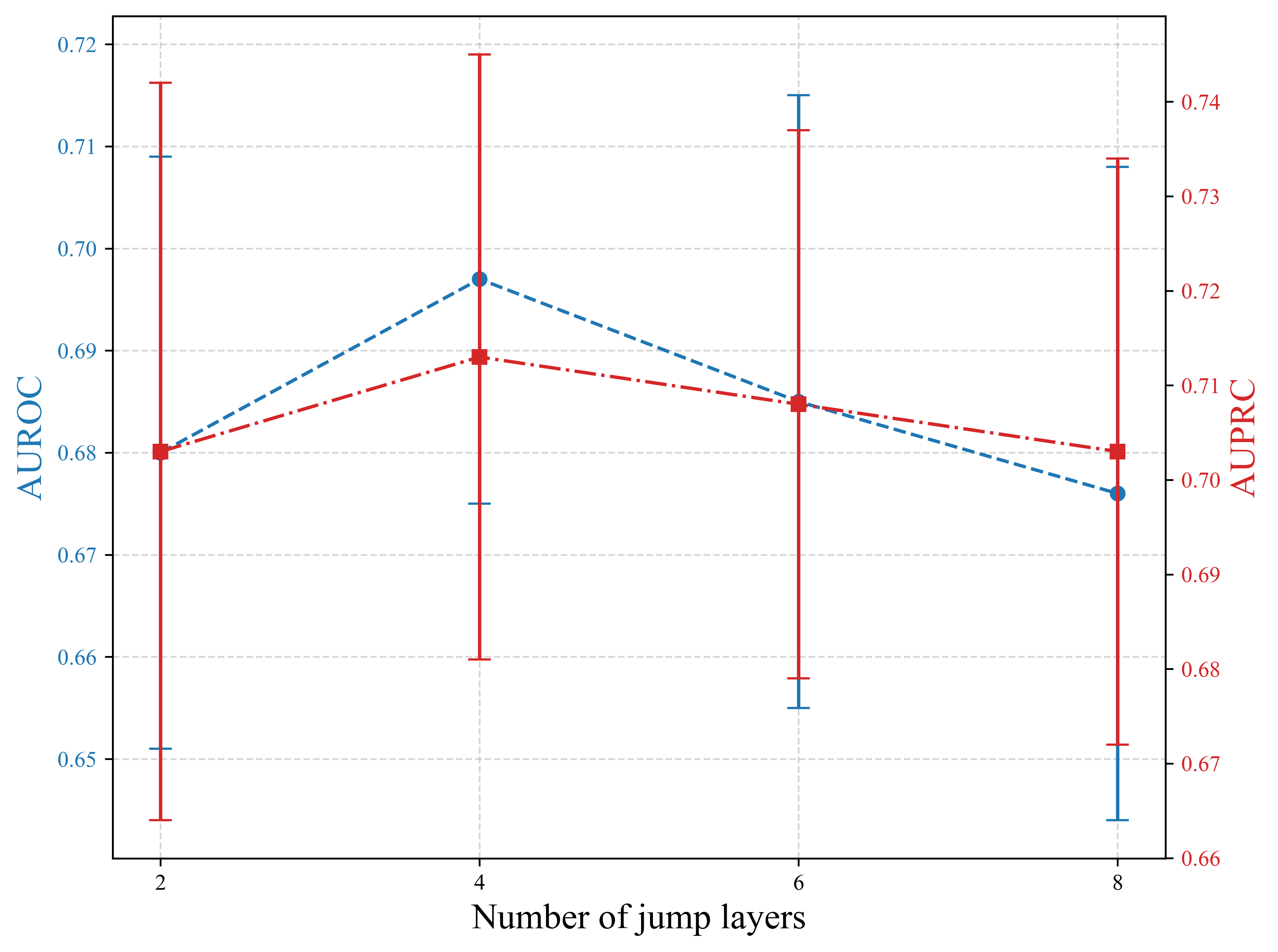}
    \caption{Sider dataset}
  \end{subfigure}
  \hfill
  \begin{subfigure}[b]{0.45\textwidth}
    \centering
    \includegraphics[width=\textwidth]{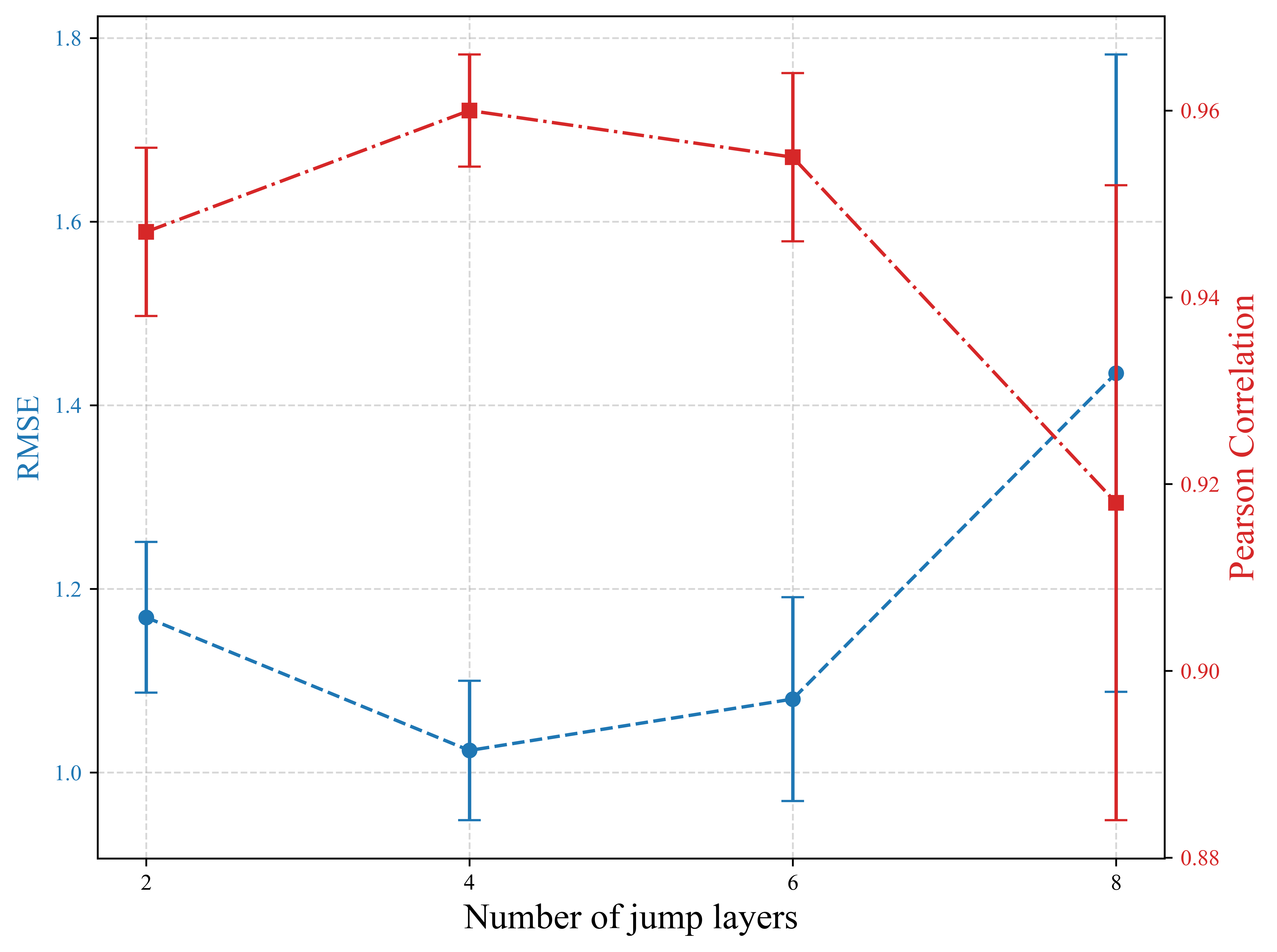}
    \caption{Freesolv dataset}
  \end{subfigure}
  \caption{\textbf{Performance of the model with varying numbers of jump layers.} (a) Results on the SIDER dataset: the red line represents AUROC, and the blue line represents AUPRC. (b) Results on the FreeSolv dataset: the red line represents RMSE, and the blue line represents Pearson correlation.}
  \label{figure:jump_layers}
\end{figure}

The results in Figure \ref{figure:jump_layers} illustrate the impact of varying the number of jump layers on the performance of MolGraph-xLSTM across the Sider and FreeSolv datasets. On the Sider dataset, the AUROC shows relatively small fluctuations, with the maximum value of 0.697 observed at 4 jump layers and the minimum value of 0.673 at 8 jump layers, representing a difference of 3.4\%. In contrast, for the FreeSolv dataset, the impact of jump layers is more pronounced. The RMSE increases significantly from its lowest value of 1.042 at 4 jump layers to its highest value of 1.326 at 8 jump layers, a difference of 27\%. The decline in performance at higher numbers of jump layers suggests that the inherent oversmoothing problem in GNNs may lead to the integration of overly smoothed deep features, which can negatively impact the performance of tasks requiring precise regression predictions.

\section{Discussion}
In this study, we propose a molecular representation learning framework that leverages xLSTM for both atom-level and motif-level graphs, providing a novel approach to molecular property prediction. Additionally, we incorporate the MHMoE module into our framework, which dynamically assigns input features to diverse expert networks, enhancing predictive accuracy through fine-grained feature activation. The effectiveness of our model is demonstrated across 10 molecular property prediction datasets, showcasing its robust performance. Additional results for other evaluation metrics are presented in the supplementary material (Table~\ref{tab:addtionalclassification} for classification tasks and Table~\ref{tab:addtionalregression} for regression tasks).

Our framework integrates atom-level and motif-level representations, and the ablation study highlights the independent effectiveness of these two levels. Specifically, both the atom-level and motif-level networks achieve competitive results individually in classification tasks (section \ref{sec4.4.1 ablation1}). However, the motif-level network exhibits a noticeable decline in regression performance. This limitation may due to the initialization features of the motif-level graph, which rely on basic substructure properties, such as the counts of specific atoms (e.g., carbon) or bond types (e.g., single bonds). While these features capture useful information for classification tasks, they may lack the precision required for accurate regression predictions. 

In addition to quantitative results, our interpretability analysis (section \ref{section: Interpretability}) highlights the strengths of the model. By analyzing the high-weight substructures identified by the model, we observed biologically meaningful correlations between the recognized substructures and specific molecular properties. This demonstrates that the model not only achieves competitive predictive performance but also provides valuable interpretability. Such interpretability is crucial for practical applications, as it can assist in drug design by guiding the identification of key molecular features associated with desired properties.

\section{Conclusion and Future Work}
Our study underscores the effectiveness of the proposed molecular representation learning framework, MolGraph-xLSTM, which leverages xLSTM for dual-level molecular graphs (atom-level and motif-level) and incorporates the MHMoE module, resulting in enhanced performance across a wide range of molecular property prediction tasks. The results of our framework demonstrate its potential for both classification and regression tasks, with notable interpretability to support applications in drug design.

However, there are areas for further improvement. The motif-level network, while effective for classification tasks, showed limitations in regression tasks, likely due to its reliance on basic substructure initialization features. Future work could focus on refining the initialization features of the motif-level network to enhance its precision in handling regression tasks. Additionally, in the atom-level branch, bond features are currently utilized in the GNN message-passing process but not in the xLSTM component. Incorporating bond-related information into the xLSTM module could further enhance the ability of the model.

Lastly, although our framework was primarily validated on molecular property prediction tasks, its versatile design as a generalizable molecular representation learning model presents opportunities for broader applications in drug discovery, including drug-target interaction prediction. Expanding into these areas could enhance the utility of the framework and drive further advancements in the field.

\bigskip
\backmatter
\bmhead{Code Availability}
The source codes for MolGraph-xLSTM are freely available on GitHub at 
\url{https://github.com/syan1992/MolGraph-xLSTM.git}.

\bmhead{Author contributions}
Conceptualization: Y.S., Y.L., Y.Y.L., Z.J., P.H.
Investigation: Y.S., Y.L., Y.Y.L., Z.J., P.H.
Data curation: Y.S., Y.L., Y.Y.L.
Formal analysis: Y.S.
Methodology development and design of methodology: Y.S., P.H.
Methodology creation of models: Y.S.
Software: Y.S.
Visualization: Y.S.
Writing original draft: Y.S.
Writing review editing: Y.S., Y.L., Y.Y.L., Z.J., P.H., C.L.
Funding acquisition: P.H.
Supervision: P.H., C.L.

\bmhead{Acknowledgements}
This work was supported in part by the Canada Research Chairs Tier II Program (CRC-2021-00482), the Canadian Institutes of Health Research (PLL 185683, PJT 190272), the Natural Sciences and Engineering Research Council of Canada (RGPIN-2021-04072), and The Canada Foundation for Innovation (CFI) John R. Evans Leaders Fund (JELF) program ($\#43481$).

\bibliography{sn-bibliography}

\end{document}


\renewcommand{\thetable}{S\arabic{table}}

\title{\parbox{\textwidth}{\centering Supplementary Information \\ MolGraph-xLSTM: A graph-based dual-level xLSTM framework with multi-head mixture-of-experts for enhanced molecular representation and interpretability}} 
\maketitle

\begin{table}[h]
\centering
\captionsetup{justification=raggedright, singlelinecheck=false}
\setlength{\abovecaptionskip}{1pt} 
\caption{Atom features and descriptions.}
\Large
\resizebox{\textwidth}{!}{
\begin{tabular}{|c|p{12cm}|c|}
\hline
\textbf{Feature Type} & \textbf{Description} & \textbf{Feature Size}  \\ \hline
\textbf{Atom Element} & The chemical symbol of the atom (e.g., C, N, O, S, etc.). & 44\\ \hline
\textbf{Atom Degree} & The number of directly bonded neighboring atoms. & 11 \\ \hline
\textbf{\# Attached Hydrogens} & The total number of hydrogen atoms directly bonded to the atom.& 11 \\ \hline
\textbf{Implicit Valence} & The implicit valence of the atom. & 11\\ \hline
\textbf{Total Valence} & The total valence of the atom. & 11\\ \hline
\textbf{Formal Charge} & The formal charge of the atom. & 11\\ \hline
\textbf{Hybridization State} & The hybridization state of the atom. & 6\\ \hline
\textbf{Radical Electrons} & The number of unpaired electrons associated with the atom. & 6\\ \hline
\textbf{Chirality} & The chirality of the atom. & 5\\ \hline
\textbf{Aromaticity} & Indicates whether the atom is aromatic. & 1\\ \hline
\textbf{Ring Membership} & Indicates whether the atom is part of a cyclic structure. & 1\\ \hline
\end{tabular}
}
\label{tab:atomfeatures}
\end{table}

\begin{table}[h]
\centering
\captionsetup{justification=raggedright, singlelinecheck=false}
\setlength{\abovecaptionskip}{1pt}
\caption{Bond features and descriptions.}
\Large
\resizebox{\textwidth}{!}{
\begin{tabular}{|c|p{12cm}|c|}
\hline
\textbf{Feature Type} & \textbf{Description} & \textbf{Feature Size} \\ \hline
\textbf{Bond Type} & The type of the bond between two atoms (e.g., single, double, triple, or aromatic). & 4 \\ \hline
\textbf{Bond Stereochemistry} & The stereochemistry of the bond. & 6 \\ \hline
\textbf{Conjugation} & Indicates whether the bond is part of a conjugated system. & 1\\ \hline
\end{tabular}
}
\label{tab:bondfeatures}
\end{table}

\begin{table}[h]
\centering
\captionsetup{justification=raggedright, singlelinecheck=false}
\setlength{\abovecaptionskip}{1pt}
\caption{Node features of motif-level graph.}
\Large
\resizebox{\textwidth}{!}{
\begin{tabular}{|c|p{10cm}|c|}
\hline
\textbf{Feature Type} & \textbf{Description} & \textbf{Feature Size} \\ \hline
\textbf{\# C} & Number of carbon atoms in the motif. & 6 \\ \hline
\textbf{\# O} & Number of oxygen atoms in the motif. & 6 \\ \hline
\textbf{\# N} & Number of nitrogen atoms in the motif. & 6 \\ \hline
\textbf{\# P} & Number of phosphorus atoms in the motif. & 6 \\ \hline
\textbf{\# S} & Number of sulfur atoms in the motif. & 6 \\ \hline
\textbf{X} & Indicates whether the motif contains a halogen atom. & 1 \\ \hline
\textbf{Other Atom} & Indicates whether the motif contains an atom other than H, C, O, N, P, S, or halogens. & 1 \\ \hline
\textbf{\# Single Bonds} & Number of single bonds in the motif. & 11 \\ \hline
\textbf{\# Double Bonds} & Number of double bonds in the motif. & 8 \\ \hline
\textbf{\# Triple Bonds} & Number of triple bonds in the motif. & 8 \\ \hline
\textbf{\# Aromatic Bonds} & Number of aromatic bonds in the motif. & 8 \\ \hline
\textbf{Ring} & Indicates whether the motif forms a ring structure. & 1 \\ \hline
\end{tabular}
}
\label{tab:motif-graph-features}
\end{table}

\begin{table}[h]
\centering
\setlength{\abovecaptionskip}{1pt}
\caption{Details of the dataset.}
\begin{threeparttable}

\begin{tabular}{|c|c|c|c|c|c|c|c|c|}
    \hline
        Dataset & \# Compounds & \# Tasks & Task Type \\ \hline
        BACE & 1513 & 1 & Binary Classification \\ \hline
        BBBP\tnote{a} & 2042 & 1 & Binary Classification \\ \hline
        HIV & 41127 & 1 & Binary Classification \\ \hline
        ClinTox & 1478 & 2 & Binary Classification \\ \hline
        Sider & 1427 & 27 & Binary Classification \\ \hline
        Tox21 & 7831 & 12 & Binary Classification \\ \hline
        Freesolv & 642 & 1 & Regression \\ \hline
        ESOL & 1128 & 1 & Regression \\ \hline
        Lipo & 4200 & 1 & Regression \\ \hline
        Caco2 & 906 & 1 & Regression \\ \hline
\end{tabular}

\begin{tablenotes}
\item[a] 11 compounds were excluded due to parsing failures with RDKit.
\end{tablenotes}
\end{threeparttable}
\label{tab:datasetdetail}
\end{table}

\begin{table}
    \centering
    \setlength{\abovecaptionskip}{1pt}
    \caption{Hyperparamter setting for each dataset.}
    \begin{tabular}{|c|c|c|c|c|c|c|c|c|}
    \hline
         & power & dimension & \#experts & \#heads & \#expert layer \\ \hline
        BACE & 4 & 256 & 8 & 16 & 2 \\ \hline
        BBBP & 4 & 256 & 8 & 8 & 2  \\ \hline
        HIV & 2 & 128 & 4 & 8 &  2  \\ \hline
        ClinTox & 4 & 128 & 8 & 8 &  1 \\ \hline
        Sider & 4 & 128 & 8 & 8 & 1   \\ \hline
        Tox21 & 4 & 128 & 8 & 8 & 2   \\ \hline
        Freesolv & 4 & 128 & 8 & 8 & 1  \\ \hline
        ESOL & 4 & 256 & 8 & 8 & 1   \\ \hline
        Lipo & 4 & 128 & 8 & 8 & 2   \\ \hline
        Caco2 & 4 & 128 & 8 & 8 & 1 \\ \hline
    \end{tabular}
    \label{tab:hyperparameter}
\end{table}

\begin{table}[]
\centering
\renewcommand{\arraystretch}{1}
\setlength{\tabcolsep}{3pt}
\resizebox{\textwidth}{!}{
\begin{minipage}{\textwidth}
\captionsetup{justification=raggedright, singlelinecheck=false}
\setlength{\abovecaptionskip}{1pt}
\caption{Additional performance evaluation on classification datasets.}
\tiny
\label{tab:addtionalclassification}
\begin{tabular}{|>{\centering\arraybackslash}m{1.4cm}  
|>{\centering\arraybackslash}m{0.8cm}|>{\centering\arraybackslash}m{0.8cm}  
|>{\centering\arraybackslash}m{0.8cm}|>{\centering\arraybackslash}m{0.8cm}
|>{\centering\arraybackslash}m{0.8cm}|>{\centering\arraybackslash}m{0.8cm}
|>{\centering\arraybackslash}m{0.8cm}|>{\centering\arraybackslash}m{0.8cm}
|>{\centering\arraybackslash}m{0.8cm}|>{\centering\arraybackslash}m{0.8cm}
|>{\centering\arraybackslash}m{0.8cm}|>{\centering\arraybackslash}m{0.8cm}|}
\hline
        & \multicolumn{2}{c|}{Sider} & \multicolumn{2}{c|}{Tox21} & \multicolumn{2}{c|}{Clintox}       
        & \multicolumn{2}{c|}{BBBP}  & \multicolumn{2}{c|}{BACE}  & \multicolumn{2}{c|}{HIV} \\ \hline
        & \multicolumn{1}{c|}{ACC} & {F1} & \multicolumn{1}{c|}{ACC} & {F1} & \multicolumn{1}{c|}{ACC} & {F1} 
        & \multicolumn{1}{c|}{ACC} & {F1} & \multicolumn{1}{c|}{ACC} & {F1} & \multicolumn{1}{c|}{ACC} & {F1}\\ \hline
FP-GNN  
& \makecell{$0.758$ \\ $\pm 0.011$} & \makecell{$0.618$ \\ $\pm 0.012$} 
& \makecell{$0.933$ \\ $\pm 0.003$} & \makecell{$0.384$ \\ $\pm 0.040$}         
& \makecell{$0.918$ \\ $\pm 0.018$} & \makecell{$0.587$ \\ $\pm 0.023$}
& \makecell{$0.841$ \\ $\pm 0.036$} & \makecell{$0.893$ \\ $\pm 0.026$} 
& \makecell{$0.768$ \\ $\pm 0.031$} & \makecell{$0.672$ \\ $\pm 0.078$}         
& \makecell{$0.972$ \\ $\pm 0.001$} & \makecell{$0.354$ \\ $\pm 0.057$}
\\ \hline
DeeperGCN 
& \makecell{$0.756$ \\ $\pm 0.015$} & \makecell{$0.605$ \\ $\pm 0.010$} 
& \makecell{$0.935$ \\ $\pm 0.001$} & \makecell{$0.290$ \\ $\pm 0.015$}         
& \makecell{$0.919$ \\ $\pm 0.032$} & \makecell{$\bm{0.689}$ \\ $\bm{\pm 0.028}$}
& \makecell{$0.835$ \\ $\pm 0.018$} & \makecell{$0.892$ \\ $\pm 0.013$} 
& \makecell{$0.772$ \\ $\pm 0.050$} & \makecell{$0.692$ \\ $\pm 0.089$}         
& \makecell{$0.969$ \\ $\pm 0.002$} & \makecell{$0.330$ \\ $\pm 0.099$}     
\\ \hline
DMPNN     
& \makecell{$\bm{0.763}$ \\ $\bm{\pm 0.016}$} & \makecell{$0.625$ \\ $\pm 0.006$} 
& \makecell{$0.936$ \\ $\pm 0.002$} & \makecell{$\bm{0.452}$ \\ $\bm{\pm 0.025}$}         
& \makecell{$0.932$ \\ $\pm 0.007$} & \makecell{$0.624$ \\ $\pm 0.076$}
& \makecell{$0.859$ \\ $\pm 0.016$} & \makecell{$0.908$ \\ $\pm 0.013$} 
& \makecell{$0.770$ \\ $\pm 0.031$} & \makecell{$0.689$ \\ $\pm 0.053$}         
& \makecell{$0.966$ \\ $\pm 0.004$} & \makecell{$\bm{0.359}$ \\ $\bm{\pm 0.057}$}   
\\ \hline
HiGNN     
& \makecell{$0.655$ \\ $\pm 0.022$} & \makecell{$0.612$ \\ $\pm 0.035$} 
& \makecell{$0.860$ \\ $\pm 0.011$} & \makecell{$0.392$ \\ $\pm 0.017$}         
& \makecell{$0.866$ \\ $\pm 0.029$} & \makecell{$0.665$ \\ $\pm 0.032$}
& \makecell{$0.825$ \\ $\pm 0.010$} & \makecell{$0.875$ \\ $\pm 0.010$} 
& \makecell{$0.781$ \\ $\pm 0.038$} & \makecell{$\bm{0.703}$ \\ $\bm{\pm 0.058}$}         
& \makecell{$0.849$ \\ $\pm 0.040$} & \makecell{$0.197$ \\ $\pm 0.052$}
\\ \hline
TranFoxMol
& \makecell{$0.756$ \\ $\pm 0.019$} & \makecell{$\bm{0.651}$ \\ $\bm{\pm 0.032}$} 
& \makecell{$\bm{0.939}$ \\ $\bm{\pm 0.003}$} & \makecell{$0.324$ \\ $\pm 0.042$}         
& \makecell{$\bm{0.940}$ \\ $\bm{\pm 0.004}$} & \makecell{$0.656$ \\ $\pm 0.024$}
& \makecell{$0.856$ \\ $\pm 0.004$} & \makecell{$0.908$ \\ $\pm 0.005$} 
& \makecell{$0.711$ \\ $\pm 0.095$} & \makecell{$0.488$ \\ $\pm 0.191$}         
& \makecell{$\bm{0.973}$ \\ $\bm{\pm 0.001}$} & \makecell{$0.153$ \\ $\pm 0.057$}
\\ \hline
MolGraph-xLSTM (Ours) 
& \makecell{$0.762$ \\ $\pm 0.012$} & \makecell{$0.647$ \\ $\pm 0.014$} 
& \makecell{$0.937$ \\ $\pm 0.002$} & \makecell{$0.421$ \\ $\pm 0.070$}         
& \makecell{$0.934$ \\ $\pm 0.002$} & \makecell{$0.647$ \\ $\pm 0.026$}
& \makecell{$\bm{0.896}$ \\ $\bm{\pm 0.020}$} & \makecell{$\bm{0.931}$ \\ $\bm{\pm 0.014}$} 
& \makecell{$\bm{0.789}$ \\ $\bm{\pm 0.033}$} & \makecell{$0.692$ \\ $\pm 0.063$}         
& \makecell{$\bm{0.973}$ \\ $\bm{\pm 0.001}$} & \makecell{$0.357$ \\ $\pm 0.083$}
\\ \hline
\end{tabular}
\end{minipage}
}

\end{table}

\begin{table}[]
\centering
\setlength{\tabcolsep}{2pt}
\resizebox{\textwidth}{!}{
\begin{minipage}{\textwidth}
\captionsetup{justification=raggedright, singlelinecheck=false}
\setlength{\abovecaptionskip}{1pt}
\caption{Additional  performance evaluation on regression datasets.}
\label{tab:addtionalregression}
\begin{tabular}{|c|cc|cc|cc|cc|}
\hline
        & \multicolumn{2}{c|}{ESOL} & \multicolumn{2}{c|}{Lipo} & \multicolumn{2}{c|}{Freesolv} & \multicolumn{2}{c|}{Caco2}  \\ \hline
        & \multicolumn{1}{c|}{MAE} & {R2} & \multicolumn{1}{c|}{MAE} & {R2} & \multicolumn{1}{c|}{MAE} & {R2} & \multicolumn{1}{c|}{MAE} & {R2} \\ \hline
FP-GNN  
& \multicolumn{1}{c|}{\makecell{$0.661$ \\ $\pm 0.014$}} & \multicolumn{1}{c|}{\makecell{$0.893$ \\ $\pm 0.012$}} 
& \multicolumn{1}{c|}{\makecell{$0.457$ \\ $\pm 0.017$}} & \multicolumn{1}{c|}{\makecell{$0.738$ \\ $\pm 0.020$}}         
& \multicolumn{1}{c|}{\makecell{$0.678$ \\ $\pm 0.159$}} & \multicolumn{1}{c|}{\makecell{$0.902$ \\ $\pm 0.046$}}
& \multicolumn{1}{c|}{\makecell{\bm{$0.360$} \\ $\bm{\pm 0.016}$}} & \multicolumn{1}{c|}{\makecell{$\bm{0.609}$ \\ $\bm{\pm 0.030}$}}
\\ \hline
DeeperGCN 
& \multicolumn{1}{c|}{\makecell{$0.470$ \\ $\pm 0.025$}} & \multicolumn{1}{c|}{\makecell{$0.906$ \\ $\pm 0.018$}}      
& \multicolumn{1}{c|}{\makecell{$0.479$ \\ $\pm 0.028$}} & \multicolumn{1}{c|}{\makecell{$0.705$ \\ $\pm 0.044$}}       
& \multicolumn{1}{c|}{\makecell{$0.692$ \\ $\pm 0.096$}} & \multicolumn{1}{c|}{\makecell{$0.878$ \\ $\pm 0.013$}}     
& \multicolumn{1}{c|}{\makecell{$0.394$ \\ $\pm 0.008$}} & \multicolumn{1}{c|}{\makecell{$0.559$ \\ $\pm 0.028$}} 
\\ \hline
DMPNN     
& \multicolumn{1}{c|}{\makecell{$0.416$ \\ $\pm 0.040$}} & \multicolumn{1}{c|}{\makecell{$0.916$ \\ $\pm 0.029$}}                 
& \multicolumn{1}{c|}{\makecell{\bm{$0.404$} \\ $\bm{\pm 0.029}$}} & \multicolumn{1}{c|}{\makecell{$0.784$ \\ $\pm 0.030$}}                 
& \multicolumn{1}{c|}{\makecell{$0.729$ \\ $\pm 0.056$}} & \multicolumn{1}{c|}{\makecell{$0.888$ \\ $\pm 0.018$}}    
& \multicolumn{1}{c|}{\makecell{$0.393$ \\ $\pm 0.010$}} & \multicolumn{1}{c|}{\makecell{$0.546$ \\ $\pm 0.031$}} 
\\ \hline
HiGNN     
& \multicolumn{1}{c|}{\makecell{$0.408$ \\ $\pm 0.041$}} & \multicolumn{1}{c|}{\makecell{$0.918$ \\ $\pm 0.024$}}                 
& \multicolumn{1}{c|}{\makecell{$0.423$ \\ $\pm 0.023$}} & \multicolumn{1}{c|}{\makecell{$0.776$ \\ $\pm 0.031$}}                 
& \multicolumn{1}{c|}{\makecell{\bm{$0.578$} \\ $\bm{\pm 0.100}$}} & \multicolumn{1}{c|}{\makecell{$0.912$ \\ $\pm 0.016$}}    
& \multicolumn{1}{c|}{\makecell{$0.373$ \\ $\pm 0.008$}} & \multicolumn{1}{c|}{\makecell{$0.584$ \\ $\pm 0.016$}}       
\\ \hline
TranFoxMol
& \multicolumn{1}{c|}{\makecell{$0.709$ \\ $\pm 0.192$}} & \multicolumn{1}{c|}{\makecell{$0.780$ \\ $\pm 0.101$}} 
& \multicolumn{1}{c|}{\makecell{$0.500$ \\ $\pm 0.042$}} & \multicolumn{1}{c|}{\makecell{$0.645$ \\ $\pm 0.088$}} 
& \multicolumn{1}{c|}{\makecell{$0.800$ \\ $\pm 0.115$}} & \multicolumn{1}{c|}{\makecell{$0.877$ \\ $\pm 0.019$}}
& \multicolumn{1}{c|}{\makecell{$0.403$ \\ $\pm 0.014$}} & \multicolumn{1}{c|}{\makecell{$0.519$ \\ $\pm 0.036$}}
\\ \hline
\makecell{MolGraph-\\xLSTM (Ours)} 
& \multicolumn{1}{c|}{\makecell{\bm{$0.385$} \\ $\bm{\pm 0.030}$}} & \multicolumn{1}{c|}{\makecell{\bm{$0.930$} \\ $\bm{\pm 0.019}$}} 
& \multicolumn{1}{c|}{\makecell{$0.412$ \\ $\pm 0.022$}} & \multicolumn{1}{c|}{\makecell{\bm{$0.787$} \\ $\bm{\pm 0.019}$}} 
& \multicolumn{1}{c|}{\makecell{$0.607$ \\ $\pm 0.071$}} & \multicolumn{1}{c|}{\makecell{\bm{$0.920$} \\ $\bm{\pm 0.014}$}} 
& \multicolumn{1}{c|}{\makecell{$0.373$ \\ $\pm 0.006$}} & \multicolumn{1}{c|}{\makecell{$0.590$ \\ $\pm 0.012$}} 
\\ \hline
\end{tabular}
\end{minipage}
}
\end{table}

\begin{figure}[ht]
    \centering
    \includegraphics[width=\textwidth]{Figure/freesolv_moe_activation.png}
    \caption{\textbf{Activation patterns of samples across the experts in the MHMoE module on the Freesolv dataset.} Each row represents a sample, and each column corresponds to an expert.}
    \label{fig:freesolv_moe_activation}
\end{figure}

\begin{figure}[h]
    \centering
    \includegraphics[width=\textwidth]{Figure/sider_moe_activation.png}
    \caption{\textbf{Activation patterns of samples across the experts in the MHMoE module on the Sider dataset.} Each row represents a sample, and each column corresponds to an expert.}
    \label{fig:sider_moe_activation}
\end{figure}